\documentclass[10pt,twocolumn,letterpaper]{article}

\usepackage{wacv}
\usepackage{times}
\usepackage{epsfig}
\usepackage{graphicx}
\usepackage{amsmath}
\usepackage{amssymb}
\usepackage{booktabs}
\usepackage{color}
\usepackage[accsupp]{axessibility}
\usepackage{pifont}
\usepackage{algorithm} 
\usepackage{algpseudocode}
\usepackage{bbm}
\DeclareMathOperator*{\argmin}{arg\,min}
\DeclareMathOperator*{\argmax}{arg\,max}
\usepackage[table]{xcolor}

\usepackage{cuted}
\usepackage{capt-of}
\usepackage{enumitem}
\usepackage{afterpage}
\usepackage{enumitem}
\setlist[itemize]{leftmargin=0.5em}

\def\wacvPaperID{446} % Enter the WACV Paper ID here

\wacvalgorithmstrack   % Uncomment this line if you are submitting to the Algorithms Track.
\wacvfinalcopy % *** Uncomment this line for the final submission

\ifwacvfinal
\usepackage{hyperref}
\hypersetup{breaklinks=true,bookmarks=false}
\else
\usepackage[pagebackref=true,breaklinks=true,colorlinks,bookmarks=false]{hyperref}
\fi

\usepackage[capitalize]{cleveref}
\crefname{section}{Sec.}{Secs.}
\Crefname{section}{Section}{Sections}
\Crefname{table}{Table}{Tables}
\crefname{table}{Tab.}{Tabs.}

\begin{document}
\setlength{\abovedisplayskip}{3pt}
\setlength{\belowdisplayskip}{3pt}

%%%%%%%%% TITLE
\title{Global-Local Self-Distillation for Visual Representation Learning}

\author{Tim Lebailly\\
KU Leuven\\
{\tt\small tim.lebailly@esat.kuleuven.be}
\and
Tinne Tuytelaars\\
KU Leuven\\
{\tt\small tinne.tuytelaars@esat.kuleuven.be}
}

\maketitle
\thispagestyle{empty}
\begin{strip}\centering
\vspace{-1.5cm}

\includegraphics[width=0.72\textwidth]{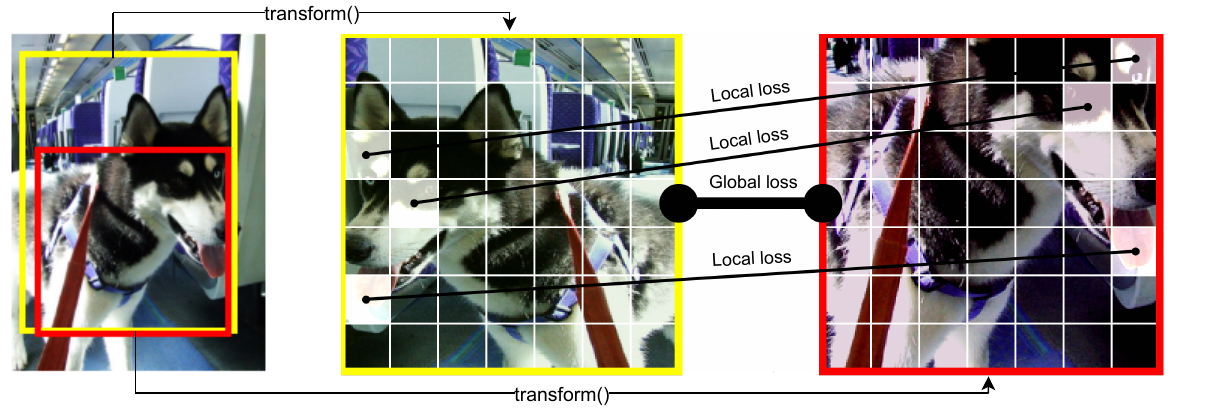}
\captionof{figure}{\textbf{Visual illustration of our global-local self-distillation framework with a loss for each component.} The global loss maximizes the similarity of both global-representations while the local losses maximize the similarity between pairs of local-representations.\label{fig:global-local_loss}}
\end{strip}

\newcommand{\TODO}[1]{{\color{red} {\bf TODO:} #1}}

\newcommand{\aug}{\tilde{\boldsymbol{x}}}
\newcommand{\im}{\boldsymbol{x}}
\newcommand{\weight}{\boldsymbol{\theta}}

\newcommand{\rep}{\boldsymbol{z}}
\newcommand{\avgp}{\bar{\boldsymbol{z}}}

\newcommand{\glob}{\bar{\boldsymbol{z}}}
\newcommand{\dense}{\boldsymbol{z}}
\newcommand{\loc}{\boldsymbol{z}^k}

\newcommand{\p}{\boldsymbol{p}}

\newcommand{\pos}{\boldsymbol{e}}

\newcommand\norm[1]{\left\lVert#1\right\rVert}

\newcommand{\augvec}{\boldsymbol{w}}

\newcommand{\augvecspa}{\boldsymbol{w}_{geo}}
\newcommand{\match}{\leftrightarrow}

\newcommand{\xmark}{\ding{55}}

% CVPR stuff
% \DeclareRobustCommand\onedot{\futurelet\@let@token\@onedot}
% \def\@onedot{\ifx\@let@token.\else.\null\fi\xspace}

% \def\eg{\emph{e.g}\onedot} \def\Eg{\emph{E.g}\onedot}
% \def\ie{\emph{i.e}\onedot} \def\Ie{\emph{I.e}\onedot}
% \def\cf{\emph{cf}\onedot} \def\Cf{\emph{Cf}\onedot}
% \def\etc{\emph{etc}\onedot} \def\vs{\emph{vs}\onedot}
% \def\wrt{w.r.t\onedot} \def\dof{d.o.f\onedot}
% \def\iid{i.i.d\onedot} \def\wolog{w.l.o.g\onedot}
% \def\etal{\emph{et al}\onedot}
\definecolor{darkgreen}{rgb}{0, 0.5, 0}

\begin{abstract}
The downstream accuracy of self-supervised methods is tightly linked to the proxy task solved during training and the quality of the gradients extracted from it. Richer and more meaningful gradients updates are key to allow self-supervised methods to learn better and in a more efficient manner. In a typical self-distillation framework, the representation of two augmented images are enforced to be coherent at the global level. Nonetheless, incorporating local cues in the proxy task can be beneficial and improve the model accuracy on downstream tasks. This leads to a dual objective in which, on the one hand, coherence between global-representations is enforced and on the other, coherence between local-representations is enforced. Unfortunately, an exact correspondence mapping between two sets of local-representations does not exist making the task of matching local-representations from one augmentation to another non-trivial. We propose to leverage the spatial information in the input images to obtain geometric matchings and compare this geometric approach against previous methods based on similarity matchings. Our study shows that not only 1) geometric matchings perform better than similarity based matchings in low-data regimes but also 2) that similarity based matchings are highly hurtful in low-data regimes compared to the vanilla baseline without local self-distillation. The code is available at \color{magenta}{\href{https://github.com/tileb1/global-local-self-distillation}{https://github.com/tileb1/global-local-self-distillation}}.

%\color{magenta}{\small{\url{https://github.com/tileb1/tim-motion-prediction}}}

% \input{fig/global-local_loss}
% show 3) similar benefits to similarity matchings in high data regimes (\eg ImageNet-1k) and should therefore be favored. 
\end{abstract}

% As such, matching local-representations from two different augmentations to obtain a local self-supervised loss is not trivial.

\section{Introduction}
\label{sec:intro}

The last few years have seen a lot a progress in self-supervised learning due to its ability to make use of large unlabeled datasets. The trend has been to train ever larger networks on ever larger datasets. However, this is very costly both in terms of compute resources and environmental impact. This also impedes research on this topic to all but a few large labs with the required infrastructure. Recent works (\eg \cite{dino,simclrv2,he2021masked}) train large models on distributed computing clusters using hundreds of GPUs for a single run. The cost for such clusters easily exceeds millions of dollars and power consumption easily surpasses the 10s of kilowatts. It is therefore crucial to make the learning as efficient as possible by leveraging as much self-supervisory signal as possible from the input images. One way to achieve this is to incorporate local cues in the self-supervised training.

Recently, transformer backbones using the self-attention mechanism \cite{all_you_need_is_attention} have been gaining more popularity in the computer vision field. Monolithic vision transformers \cite{vit16x16} have induced a wave of work on multi-stage vision transformers \cite{swin,vil,cvt,ranftl2021vision} which do not process patch tokens at a single resolution (\eg 16x16) but at multiple resolutions via patch merging. These architectures encode an input image into a representation which is coarser-grained than the pixel-level, yet preserves the spatial structure of the input image. Given the highly complex mapping from input image to output representation, local regularization is even more motivated. 

Typical self-distillation frameworks aim to maximize the similarity between the global representation of two augmented crops coming from the same input image. The idea is to generate augmentations which carry the same semantic meaning (\eg image of dog) but contain different low-level information (\eg different lighting, background, scale \etc). The backbone is then trained to output a representation of both augmented images which are coherent with each other. Under this setup, the network should learn to retain semantic content from the input image while discarding redundancy and noise. To incorporate additional self-supervisory signal in the training, one can devise a similar loss which acts on local-representations instead of the global-representation. This leads to a dual objective where two terms are optimized (local and global) as shown in \Cref{fig:global-local_loss}. As opposed to the global-representation, a single augmented image does not lead to a single local-representation but to a set of local-representations. This makes the expression for the local loss non-trivial. A question naturally arise: \textit{How should one generate pairs of local-representations \ie which local-representation from one augmented image should be matched to which local-representation from the other augmented image?}

Ideally, we would like to match local-representations which share the same semantic content. In a self-supervised setup, we don't have access to such oracle and should rely purely on data-driven approximations. Li \etal \cite{esvit} propose to use a matching function which is based on the similarity of local-representations. The assumption is that similar local-representations should be semantically close. In practice, this does not always hold, especially when augmented images don't overlap much. On the other hand, we propose a geometric matching function. The assumption is that representations originating from close-by regions of the input image are semantically close. One can also easily threshold the matching distance to avoid the above mentioned problem of little overlap between augmentations. We propose a study comparing both approaches and summarize our contribution as follows:
\vspace{-0.1cm}
\begin{itemize}
    \setlength\itemsep{-0.1em}
    \item To our knowledge, we are the first to introduce local self-distillation for the features of multi-stage vision transformers based on geometry.
    \item We study what is the best way to incorporate local self-distillation including a similarity based proxy task as in SOTA method \cite{esvit} and our geometry based self-supervised proxy task:
    \vspace{-0.1cm}
    \begin{itemize}
    \setlength\itemsep{0em}
        \item We show that a similarity based self-distillation proxy task can be hurtful in low-data regimes and performs much worse than the vanilla setup without additional local loss. The geometry based self-distillation proxy task is more robust and improves the vanilla setup in all data-regimes.
        \item We show comparable performance between both approaches in high-data regimes (\eg ImageNet-1k \cite{imagenet}).
        \item Finally, we show that geometry based matchings lead to a processing of the input image which better preserves the spatial structure of images and show empirical evidence of local-representation mode collapse when using a similarity based matching function.
    \end{itemize}
\end{itemize}

\section{Related works}
%These tasks are synthetic problems for which no labels are required for learning the weights of an encoder.

%These are just a few examples of the numerous pretext tasks studied in the literature. 
% We review four closely related topics \ie \textit{self-supervised learning}, \textit{dense self-supervised learning}, \textit{vision transformers} and \textit{positional encodings}.\\%, as well as the major works in each topic.

%\subsection{Self-supervised learning}
\noindent
{\bf Self-supervised learning} Early self-supervised methods for representation learning make use of pretext tasks. As pretext tasks, Noroozi et al.~\cite{noroozi_jigsaw} solve a jigsaw puzzle and Gidaris et al.~\cite{gidaris_rotations} predict which rotation was applied on an input image. Other approaches include predicting patch context \cite{contex_pred_doersch,contex_pred_Mundhenk}, inpainting patches \cite{inpainting}, predicting noise \cite{predicting_noise}, \etc.

%The representations of similar images are pulled together in embedding space and the representations of highly distinct images are pushed away from each other.
More recently, contrastive learning methods have been the most popular. Contrastive learning is a scheme for metric learning which leverages distinctiveness and similarities between inputs. Chen \etal~\cite{pmlr-v119-chen20j} propose a simple framework for contrastive learning of visual representations (SimCLR) which has kickstarted a lot of research in this direction \cite{simsiam,moco,simclrv2,mocov2,mocov3,moby,DBLP:journals/corr/abs-2102-08318,DBLP:journals/corr/abs-2111-12309}. Grill \etal propose a framework called BYOL \cite{byol} where negative samples in the contrastive loss are not needed to avoid collapse by using a simple mean squared error loss (MSE) between the output representations of two branches. Caron \etal \cite{dino} (DINO) extend this framework by introducing visual transformers as the backbone and by viewing this learning paradigm as self-distillation. We were inspired by DINO, yet observed that they only use the global representations, leaving valuable cues at the local scale unexploited. Note that self-supervised methods require large quantities of data to get great results. Few works focus explicitly on self-supervised pretraining on small-scale datasets \cite{ciga2021resource,saillard2021self,liu2021efficient}.\\[-10pt] 
%Some of SimCLR's contribution includes the addition of a learnable non-linear transformation between the representation and the contrastive loss as well as the empirical observation that stronger augmentations lead to better accuracy on downstream tasks. 

%These representations can be used to solve downstream tasks like classification. 
%These representations can be used to solve downstream tasks like classification. 
\noindent
{\bf Dense contrastive learning}
The above-mentioned methods focus on learning a visual representation at the image-level. Some works take a different approach and aim to learn a representation at the pixel-level, which is useful for dense tasks like segmentation. Pinheiro \etal \cite{unsupervised_learning_of_dense_visual_rep} generate positive pixel pairs corresponding to the same location from an input image. Xie \etal \cite{pixpro} use a similar loss as well as an additional pixel-to-propagation consistency task improving the downstream task accuracy. Notable works along the same lines include \cite{DBLP:journals/corr/abs-2103-04814,Wang_2021_CVPR}.\\
%Wang \etal \cite{Wang_2021_CVPR} do not use a pixel-matching scheme based on location in the input image but based on similarity of the pixel-representations. 
\noindent
{\bf Dense self-distillation}%Closer to our work, 
$\text{ }$Li \etal (EsViT \cite{esvit}) focus on classification downstream tasks and propose a self-distillation task leveraging the local features of a multi-stage visual transformer (rather than pixel-representations) based on their similarities. EsViT is thus not fully dense but does share similarities with these approaches. We argue that explicitly using the spatial information from the original input images, as we do, provides stronger feedback than matching local representations purely based on similarity, as they do (especially in low-data regimes). \\[-10pt] %However, finding local-representations matchings based on spatial location is not as easy as for the pixel-representations since the data augmentation pipeline is such that it generates local-representations which are not aligned on the same grid.%\footnote{We tried adapting the data-augmentation pipeline to only generate augmentations which leave the local-representations aligned on the same grid by using fixed re. However, }.

%Our work was developed largely in parallel to ESViT.

% \subsection{Small scale self-supervised learning}
% \TODO{add section on low data}
\noindent
{\bf Vision transformers}
Visual transformers (ViT) have been proposed by Dosovitskiy \etal \cite{vit16x16} as an alternative to the more common CNN backbone (\eg ResNets \cite{resnet}). Input images are patchified, then each patch is flattened and fed to a linear layer whose output serves as tokens in a traditional NLP transformer backbone \cite{all_you_need_is_attention}. ViTs allow more complex mappings between input and output as the architecture is not translation invariant. In the absence of a supervisory training signal, ViTs are adequate to model complex dependencies and outperform ResNets, as shown by DINO \cite{dino}. Some self-supervised works mimic the \textit{masked word prediction} task in NLP with a \textit{masked image modeling} task \cite{masked_image_modeling,mae}. More recently, multi-stage architectures have been proposed where patches are not processed at a single resolution but at multiple resolutions via patch merging. Liu \etal~\cite{swin} propose such an architecture where they also process tokens in windows of restricted size to lower the compute requirements. Other notable works along the same lines include \cite{focal_transformer,vil,cvt,ranftl2021vision,relative_positional_encoding,conditional_posenc}.\\[-10pt] 

\begin{figure}[t]
  \centering
   \includegraphics[width=\linewidth]{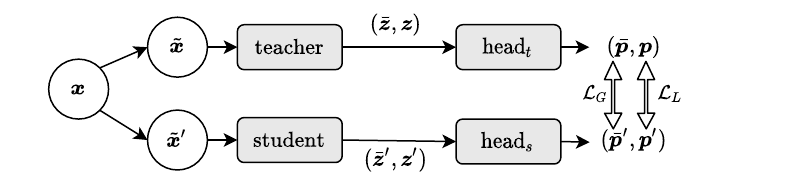}

   \caption{\textbf{High-level overview of a global-local self-distillation framework.} The augmentations $\aug$ and $\aug'$ are fed respectively through the teacher and student backbone resulting in both global- ($\glob$) and dense-representations ($\dense$). Then, these representations are respectively fed through a teacher- and student-head to output probability mass functions with which $\mathcal{L}_G$ and $\mathcal{L}_L$ are minimized (more details in \cref{sec:loss}).}
   \label{fig:sketch}
\end{figure}

\section{Methodology}
\label{sec:methodology}

This section starts by reviewing necessary representation terminology (\cref{sec:globalvslocal}) and the augmentation pipeline (\cref{sec:augmentation_pipeline}). Then, the training scheme is discussed (\cref{sec:knowledge-distilation}). Finally, we review the self-supervised loss needed to incorporate additional local cues. A high-level overview sketch can be found in Figure \ref{fig:sketch}.% where the local loss $\mathcal{L}_L$ takes a different form depending on how the local cues are incorporated.

\subsection{Global- versus local-representations}
\label{sec:globalvslocal}
Most previous works (\eg \cite{dino,pmlr-v119-chen20j,byol,simsiam,moco}) use a self-supervised loss based only on the \textit{global-representation} of the augmentations. With this term, we refer to either the output of the backbone network after a global average pooling or the \texttt{[CLS]} token in the case of vision transformer backbones. In both cases, the global-representation is a vector $\glob \in \mathbb{R}^d$ where $d$ is the size of the latent space. On the other hand, we use the term \textit{dense-representation} to refer to a representation in which spatial structure of the input image is explicitly modeled. Such dense-representations usually take the form of a third-order tensor $\in \mathbb{R}^{H\times W \times d}$ or $\in \mathbb{R}^{HW \times d}$, where $H$ and $W$ are respectively the height and width of the input image or a downscaled version thereof. Examples of such dense-representation include the output feature map of a CNN or an ordered sequence of tokens from a visual transformer (excluding the \texttt{[CLS]} token). Finally, we use the term \textit{local-representation} to refer to a 1D slice $\rep_k \in \mathbb{R}^d$ of a dense-representation associated to a certain local position $k$ (of the $K=HW$ possible locations).

% Incorporating local cues in a traditional self-supervised framework leads to an objective with two terms acting independently on the global- and local-representations (\cref{sec:similarity_local_loss}).

\subsection{Data augmentation pipeline as a composition of geometric and photometric transforms}
\label{sec:augmentation_pipeline}
There are 3 main components in self-supervised frameworks: 1) a data-augmentation pipeline, 2) a backbone and 3) a self-supervised proxy task. Data augmentation pipelines play a crucial role in self-supervised learning settings since they produce the necessary augmented samples needed to enforce a self-supervised loss. Previous work \cite{whatmakesforgoodviews} shows empirical findings on how the downstream task accuracy is linked to parameters of the pipeline. In our work, we assume the data augmentation pipeline as given and use the same one as \cite{dino} and \cite {esvit}. This pipeline is the fruit of empirical testing from many previous works \cite{pmlr-v119-chen20j,hjelm2019learning,NEURIPS2019_ddf35421,swav,dino}.

The data augmentation pipeline is a long composition of multiple transforms, including both geometric transforms and photometric transforms. Geometric transforms include \texttt{CROP}, \texttt{RESIZE} and \texttt{HORIZONTAL\_FLIP} while photometric transforms include \texttt{COLOR\_JITTER}, \texttt{SOLARIZE}, \texttt{GAUSSIAN\_BLUR} and \texttt{GRAYSCALE}. We denote the composition of all geometric transforms by $\mathbf{G}$ and the composition of all photometric transforms by $\mathbf{P}$.

Geometric and photometric transforms are respectively parametrized by vectors $\augvec_{geo}$ and $\augvec_{pho}$ with $\augvec_{geo} = [ul_x, ul_y, lr_x, lr_y, h, w, f]$. The first 4 elements represent the location of the crop (in the form of upper left and lower right coordinates) to be taken w.r.t. the original image. $h$ and $w$ refer to the resized shape of the crop while $f$ is a binary variable indicating whether the crop is flipped horizontally or not. The actual form of $\augvec_{pho}$ is not relevant for this analysis. The data augmentation pipeline is characterized by a distribution $\mathcal{D}_{aug}$ from which all parameters are sampled, \ie $\augvec \sim \mathcal{D}_{aug}$ with $\augvec=[\augvec_{geo},\augvec_{pho}]$.

Given a single input image $\im$ and a sampled augmentation parameter vector $\augvec$, we generate an augmentation $\aug$ as follows
\begin{equation}
    \aug = \mathbf{P}\left(\mathbf{G}(\im, \augvec_{geo}), \augvec_{pho}\right)
\end{equation}

% Traditionally, self-supervised losses are only functions of two augmentations $\aug$ and $\aug'$, \ie $\mathcal{L}_{SSL}(\aug, \aug')$. To incorporate additional local cues in a geometric fashion, these functions also need to have access to the geometric augmentation vectors $\augvec_{geo}$ \ie $\mathcal{L}_{SSL}(\aug, \aug', \augvec_{geo}, \augvec_{geo}')$.
\begin{figure*}[bt]
  \centering
  \includegraphics[width=\linewidth]{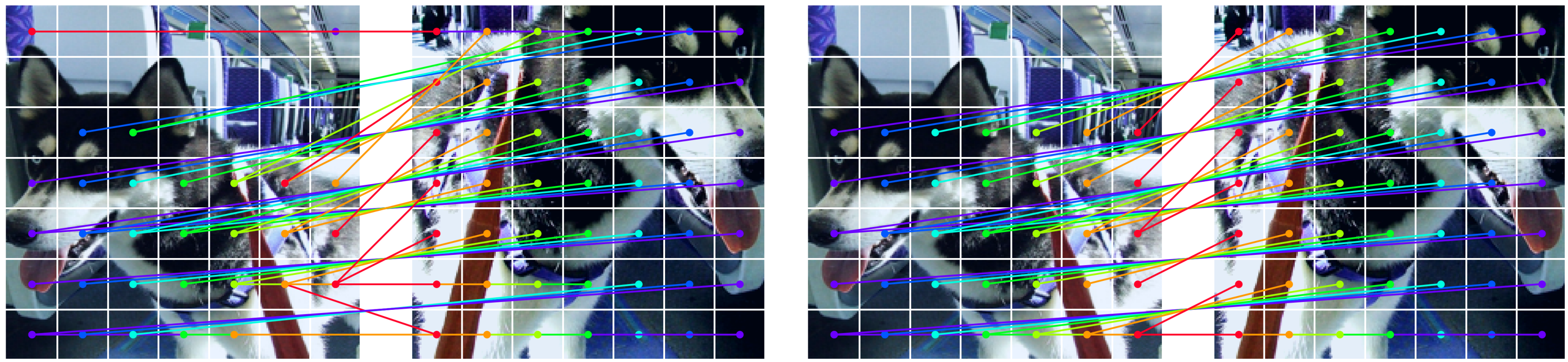}
  \caption{\textbf{Visualization of the matchings enforced during training for $\mathcal{L}_L^{sim}$ (left) and $\mathcal{L}_L^{geo}$ (right).} The similarity matchings depend on the state of the backbone during the training phase as they are a function of the local-representations (here shown after 300 epochs of training on ImageNet-1k). The geometric matchings on the other hand are not a function of the local-representations and hence are fixed throughout the training. Colors are used only to better distinguish different matchings.}
  \label{fig:comparison_matchings}
  \vspace{-0.4cm}
\end{figure*}

% \TODO{turn into 1 fig}
% \begin{figure}
%     \centering
%     \begin{minipage}{0.5\textwidth}
%         \centering
%         \includegraphics[width=0.9\textwidth]{fig/similarity.png} % first figure itself
%         \caption{\textbf{Visualization of the matchings enforced during training for $\mathcal{L}_L^{sim}$}. These matchings depend on the state of the backbone during the training phase as they are a function of the local-representations (here shown after 300 epochs of training on ImageNet-1k).}
%         \label{fig:similarity_training_matchings}
%     \end{minipage}\hfill
%     \begin{minipage}{0.5\textwidth}
%         \centering
%         \includegraphics[width=0.9\textwidth]{fig/geometric.png} % second figure itself
%         \caption{\textbf{Visualization of the matchings enforced during training for $\mathcal{L}_L^{geo}$}. These matchings are not a function of the local-representations and hence are fixed throughout the training.}
%         \label{fig:geometric_training_matchings}
%     \end{minipage}
% \end{figure}

\subsection{Self-distillation}
\label{sec:knowledge-distilation}
Before we can dive into the explicit expression of the loss, we first review the self-supervised training scheme which we use in our work. Within self-supervised learning methods, contrastive ones are the most popular and have been used mainly due to their ability to avoid mode collapse in an explicit and simple manner. However, \cite{simsiam,dino,byol} show that negative samples are not needed to learn representations while avoiding collapse by including some of the following tricks: use asymmetric predictors, use stop-gradients in one branch, have one branch reflect a low-passed (\eg with an exponential moving average) version of the other, run more gradient descent steps on one branch, use some kind of normalization on the representation, \etc. However, \cite{dino} are the first to view ``contrastive learning without negative pairs'' as a form of self-supervised knowledge distillation. Knowledge distillation is a learning paradigm where a student network learns to imitate the output of a teacher network. In a self-supervised setting, both the student $g_{s}$ and the teacher network $g_{t}$, parametrized respectively by $\weight_s$ and $\weight_t$, are initialized to the same random $\weight_{init}$. The student is then optimized such that its output matches the one of the teacher w.r.t. a particular loss function. The teacher network is updated at each epoch to reflect an exponential moving average of the student's weights, \ie $\weight_t \leftarrow \lambda \weight_t + (1-\lambda) \weight_s$.

%%%%%%%%%%%%%%%%%%%%%%%%%%%%%%%%%%%%%%%%%%%%%%%%%%%%%%%%%%%%%%%%%%%%%%%%%%%%%%%%%%%%%%%%%%%%%%%%%%%%%%%%%%%%%%
\subsection{Self-supervised loss}
\label{sec:loss}
The global-representation loss used in our study is the same as proposed by DINO~\cite{dino} and is explained in the following subsection using notations similar to EsViT~\cite{esvit}. Given a backbone $f$ and an augmentation $\aug$, we obtain both the global- ($\glob$) and the dense-representation ($\dense$) in a single forward pass, \ie $(\glob, \dense) = f(\aug)$. By abuse of notations, we will use $\glob = \bar{f}(\aug)$ and $\dense = f(\aug)$.

Given a student backbone $f_s$ and teacher backbone $f_t$ as well as a set $\mathcal{V} = \{\aug^1, \aug^2, \aug^3, \cdots\}$ containing $N=|\mathcal{V}|$ augmented views of the same input image, a single forward pass of all augmentations in both networks results in:
\begin{enumerate}
    \item two sets of global-representations $\bar{\mathcal{Z}}_s = \{\bar{f_s}(\aug) : \aug \in \mathcal{V}\}$ and $\bar{\mathcal{Z}}_t = \{\bar{f_t}(\aug) : \aug \in \mathcal{V}\}$
    \item two sets of local-representations ${\mathcal{Z}}_s = \{{f_s}(\aug) : \aug \in \mathcal{V}\}$ and ${\mathcal{Z}}_t = \{{f_t}(\aug) : \aug \in \mathcal{V}\}$
\end{enumerate}

%%%%%%%%%%%%%%%%%%%%%%%%%%%%%%%%%%%%%%%%%%%%%%%%%%%%%%%%%%%%%%%%%%%%%%%%%%%%%%%%%%%%%%%%%%%%%%%%%%%%%%%%%%%%%%
\subsubsection{Global-representation loss}
\label{sec:global_loss}
% \cref{alg:savit}
The global-representations $\glob$ are then mapped to a discrete probability mass function $\bar{\p}$ of dimension $I$ using an MLP-head $\bar{h}$, \ie $\bar{\p} = \bar{h}(\glob)$. For each pair of global-representations coming from the student and the teacher, we use $\bar{h}$ to map them to a probability mass function and minimize their cross-entropy, more explicitly\footnote{We choose to leave \cref{eq:global_loss} in a more readable format avoiding the multi-crop strategy \cite{swav} which we do use in practice. The full expression for the loss with the multi-crop strategy can be found in the appendix.}:
%
%Let us denote $\bar{h} \circ \bar{f}$ using $\bar{g}$
%
% \begin{equation}
%     \mathcal{L}_G = \frac{1}{|\mathcal{V}|(|\mathcal{V}|-1)}\sum_{\aug \in \mathcal{V}} \sum_{\substack{\aug' \in \mathcal{V} \\ \aug \neq \aug'}} H(\bar{g_t}(\aug), \bar{g_s}(\aug'))
%     \label{eq:global_loss}
% \end{equation}
%
\begin{equation}
    \mathcal{L}_G = \frac{1}{N(N-1)} \sum_{\glob \in \bar{\mathcal{Z}}_t} \sum_{\substack{\glob' \in \bar{\mathcal{Z}}_s \\ \aug \neq \aug'}} H\left(\bar{h}(\glob), \bar{h}(\glob')\right)
    \label{eq:global_loss}
\end{equation}
with
\begin{equation}
    H(p, q) = -\sum_{i \in \mathcal{I}} p(i) \log q(i)
\end{equation}
where $\mathcal{I}$ is the support of the distributions $p$ and $q$, in our case $\mathcal{I}=[I]=\{1, 2, \cdots, I\}$. The summation constraint $\aug \neq \aug'$ of the inner sum refers to the fact that we do not have a term $H\left(\bar{h}(\glob), \bar{h}(\glob')\right)$ where $\glob$ and $\glob'$ are global-representations corresponding to the same augmentation.

%%%%%%%%%%%%%%%%%%%%%%%%%%%%%%%%%%%%%%%%%%%%%%%%%%%%%%%%%%%%%%%%%%%%%%%%%%%%%%%%%%%%%%%%%%%%%%%%%%%%%%%%%%%%%%
\subsubsection{Similarity based local-representation loss}
\label{sec:similarity_local_loss}
Similar to the global-representation loss, each local-representation $\dense_k,  \forall k \in [K]$ is mapped to a probability mass function $\p_k$ using another MLP head $h$, \ie $\p_k = h(\dense_k)$. A local-representation $\dense_k$ from an augmented $\aug$ is matched to the best corresponding $\dense'_{k^\star}$ from another augmented image $\aug'$. Here, the best corresponding local-representation is selected as the local-representation $\dense'_{k^\star}$ in the other augmentation $\aug'$ which has the highest similarity with $\dense_{k}$ from the first augmentation $\aug$ as done in EsViT~\cite{esvit}. This is shown on the left of \Cref{fig:comparison_matchings}. The cross-entropy between the probability outputs of matching local-representations is then minimized for all matchings and all pairs of augmentations $\aug$ and $\aug'$. Given two dense representations $\dense$ and $\dense'$:

\begin{equation}
    L_L^{sim}(\dense, \dense') = \frac{1}{K} \sum_{k \in [K]} H\left(h(\rep_k), h(\rep'_{k^\star})\right)
    \label{eq:similarity_local_loss}
\end{equation}

where $k^\star = \argmax_j \frac{\dense_k^\top \dense'_j}{\norm{\dense\vphantom{'}_k}\norm{\dense'_j}}$. Averaging over all pairs of dense representations, the total local similarity based self-supervised objective becomes
\begin{equation}
    \mathcal{L}_{L}^{sim} = \frac{1}{N(N-1)} \sum_{\rep \in \mathcal{Z}_t} \sum_{\substack{\rep' \in \mathcal{Z}_s \\ \aug \neq \aug'}} L_L^{sim}(\dense, \dense')
    \label{eq:total_similarity_loss}
\end{equation}

\subsubsection{Geometric local-representation loss}
\label{sec:geometric_local_loss}
Along the set of augmented views $\mathcal{V}$, we also dispose over a set $\mathcal{W}_{geo} = \{\augvecspa^1, \augvecspa^2, \augvecspa^3, \cdots\}$ of vectors $\augvecspa$ which describe the geometric transforms $\im$ has undergone to generate $\aug$. Using this set, we generate another set $\mathcal{E} = \{\pos^1, \pos^2, \pos^3, \cdots\}$ where each element $\pos \in \mathbb{R}^{H \times W \times 2}$ is an object of the same spatial dimension as its associated dense-representation $\dense \in \mathbb{R}^{H \times W \times d}$. $\pos_k \in \mathbb{R}^2$ (a slice of $\pos$) encodes the $(x,y)$ coordinates of the center point of the patch associated to $\dense_k$ for every $K=HW$ locations in $\dense$ w.r.t. the original input image grid $\im$ (not the augmentation $\aug$). Note that $\pos$ is a ``positional encoding'', though it should not be confused with the positional encoding used in transformers to remove the permutation invariance of the tokens.

Here, the best corresponding $\dense'_{k^\star}$ from another augmented image $\aug'$ is selected based on how close they are w.r.t. to the original image grid $\im$. The cross-entropy between the probability outputs of matching local-representations is then minimized for most matchings and all pairs of augmentations $\aug$ and $\aug'$. As opposed to the similarity based local loss, we do not average over all pairs of local-representations. A matching $\dense_k \match \dense'_{k^\star}$ obtained via $k^\star = \argmin_j \norm{\pos_k - \pos'_j}^2$ might be very bad (in terms of matching distance $=d(k)=\min_j \norm{\pos_k - \pos'_j}$) when there is no overlap between $\aug$ and $\aug'$. Therefore, we restrict our averaging over the set of matchings $\dense_k \match \dense'_{k^\star}$ which have a low matching distance \ie the $\dense_k$ and $\dense'_{k^\star}$ lie on the region of overlap between $\aug$ and $\aug'$. The matching distance threshold $s$ is set to half of the maximum between 1) the length of the diagonal of a local representation $\dense_k$ corresponding to augmentation $\aug$ and 2) the length of the diagonal of a local representation $\dense'_k$ corresponding to augmentation $\aug'$. By the length of the diagonal of a local representation $\dense_k$, we refer to the Euclidean distance between $\pos_k$ and an adjacent diagonal $\pos_{k^*}$ where $\pos$ is the positional encoding described in the first paragraph of \cref{sec:geometric_local_loss}. s is set to this value because if $d(k) > s$, either $\dense_k$ or $\dense_{k^\star}'$ falls outside the region of overlap (if any) between  augmentations $\aug$ and $\aug'$. Taking the above into consideration and given two dense representations $\dense$ and $\dense'$:

\begin{equation}
    L_L^{geo}(\dense, \dense') = \frac{1}{K} \sum_{k \in [K]} \mathbbm{1}_{\{d(k)<s\}} H\left(h(\rep_k), h(\rep'_{k^\star})\right)
    \label{eq:geometric_local_loss}
\end{equation}
%
% \begin{equation}
%     \mathcal{L}_L = \frac{1}{N(N-1)} \sum_{\rep \in \mathcal{Z}_t} \sum_{\substack{\rep' \in \mathcal{Z}_s \\ \aug \neq \aug'}} \frac{1}{K} \sum_{k \in [K]} H\left(h(\rep_k), h(\rep'_{k^\star})\right)
%     \label{eq:local_loss}
% \end{equation}
%
where $k^\star = \argmin_j \norm{\pos_k - \pos'_j}^2$. $\pos$ and $\pos'$ are the positional encodings associated respectively to $\dense$ and $\dense'$. $d(k)=\min_j \norm{\pos_k - \pos'_j}$ is the matching distance of $\dense_k \match \dense'_{k^\star}$, $s$ is the dynamically set distance threshold and $\mathbbm{1}_{\{\text{condition}\}}$ is the indicator function. Averaging over all pairs of dense representations, the total local self-supervised objective based on geometry becomes

\begin{equation}
    \mathcal{L}_{L}^{geo} = \frac{1}{N(N-1)} \sum_{\rep \in \mathcal{Z}_t} \sum_{\substack{\rep' \in \mathcal{Z}_s \\ \aug \neq \aug'}} L_L^{geo}(\dense, \dense')
    \label{eq:total_geometric_loss}
\end{equation}

%%%%%%%%%%%%%%%%%%%%%%%%%%%%%%%%%%%%%%%%%%%%%%%%%%%%%%%%%%%%%%%%%%%%%%%%%%%%%%%%%%%%%%%%%%%%%%%%%%%%%%%%%%%%%%
%

In the following section, we study the effect of the additional local cues by varying the total self-supervised objective in three different settings: \texttt{Vanilla}, \texttt{Similarity} and \texttt{Geometric}. An overview of the three settings can be found in \Cref{tab:rough_comparison}. The sum of the global- and local-loss is the total objective which is optimized w.r.t. the parameters of the student network. Pseudo code for our the \texttt{Similarity} and \texttt{Geometric} setting can be found in the appendix.

\begin{table}[t]
\caption{\textbf{Overview of three different settings based on the local loss used.} $\mathcal{L}_G$ and $\mathcal{L}_L^{sim/geo}$ refer respectively to the global- and local-representation loss. The matching type column refers to the matching type in the local-representation loss.}
\vspace{0.2cm}
\resizebox{\columnwidth}{!}{
\begin{tabular}{l|c|c|c|c|c}
\hline
\textbf{Setting} & \textbf{Backbone} & \textbf{Global loss} & \textbf{Local loss}    & \textbf{Matching type} & \textbf{Proposed in} \\
\hline
\texttt{Vanilla} & Swin-T/7      & $\mathcal{L}_G$ &          \xmark     &  \xmark                   & DINO \cite{dino}\footnotemark     \\
\texttt{Similarity} & Swin-T/7     & $\mathcal{L}_G$ & $\mathcal{L}_{L}^{sim}$ & similarity          & EsViT \cite{esvit}     \\
\texttt{Geometric} & Swin-T/7     & $\mathcal{L}_G$ & $\mathcal{L}_{L}^{geo}$ & geometry   & New (ours)    
\end{tabular}
}
\label{tab:rough_comparison}
\end{table}
\footnotetext{Note that we replace the ViT backbone with a Swin transformer so that the only difference between the three settings is the local loss.}

\subsubsection{Computational complexity of the local loss}
The local loss leads to limited compute overhead since it only adds an additive term in both the forward and backward pass which is very small compared to the backbone computations. In the \texttt{Similarity} setting (with local loss $\mathcal{L}_L^{sim}$), given two dense-representations $\dense, \dense' \in \mathbb{R}^{HW \times d}$, the compute complexity is $O(H^2W^2d)$. This is for $(HW)^2$ inner products, each of cost $O(d)$. The argmax operation is only $O(HW)$. In the \texttt{Geometric} setting (with local loss $\mathcal{L}_L^{geo}$), given two positional encodings $\pos$ and $\pos' \in \mathbb{R}^{HW \times 2}$, the compute complexity is $O(H^2W^2)$. This is for $(HW)^2$ L2-norms, each of cost $O(1)$. The argmin operation is only $O(HW)$. With a Swin-T backbone and 224x224 input images, $H=W=7$ and $d=192$ this results in a negligible cost compared to the computations in the backbone.

%%%%%%%%%%%%%%%%%%%%%%%%%%%%%%%%%%%%%%%%%%%%%%%%%%%%%%%%%%%%%%%%%%%%%%%%%%%%%%%%%%%%%%%%%%%%%%%%%%%%%%%%%%%%%%

\section{Results}
% In a first step, we evaluate all methods using ImageNet-1k \cite{imagenet} as pretraining dataset and compare it against methods using backbones of similar size. Next, we show the performance benefits of using local cues when pretraining in low-data regimes. Finally, we analyze the correspondence matchings computed based on their cosine similarity. The key takeaway from this section is that the addition of local cues in the self-supervised loss is beneficial in most settings when using geometry based matchings which is not the case for similarity based matchings.

\subsection{Rationale of the experiment design}
To evaluate the merits of the additional geometric local self-distillation, we compare the downstream performance of this method with the \texttt{Vanilla} and \texttt{Similarity} settings in which the local loss is removed or the geometric local loss is replaced by a similarity local loss (SOTA method). We compare the 3 representation learning methods on ImageNet-1k using the linear and k-NN benchmarks which are industry standard evaluations (\cref{sec:imagenet1k}). To get a grasp of the robustness of the methods depending on the dataset size, we run these benchmarks on randomly sampled subsets of ImageNet-1k. We observe an improvement of our method in all data regimes as well as a large performance drop for the \texttt{Similarity} setting (SOTA method). To corroborate our results in low-data regimes, we run the same study on smaller scale datasets (as well as multiple different backbones) in which analogous conclusions can be drawn (\cref{sec:otherdatasets}). We hypothesise that the large performance drop of the \texttt{Similarity} setting can be due to a collapse at the local level and show empirical evidence to confirm that (\cref{sec:collapse}). Additionally, we propose a correspondence matching analysis (both qualitative and quantitative) to observe the effect of the local losses (\cref{sec:correspondence_matching}).

We mostly focus on classification downstream tasks as opposed to dense tasks \eg object detection. Dense evaluations are usually solved by fine tuning Mask-RCNN \cite{maskrcnn} on top of the pretrained backbone. As such it is hard to distinguish whether a high downstream accuracy is due to a good pretraining or due to the added capacity of Mask-RCNN. Recent work (see Table 1 of \cite{https://doi.org/10.48550/arxiv.2111.11429}) even shows better downstream accuracy on a randomly initialized network than on a pretrained one with MoCo v3. \textbf{k-NN and linear evaluation bechmarks for classification are better candidates to evaluate the intrinsic quality of the pretraining} since they don't require much processing. We do evaluate a dense downstream task with little processing in \Cref{sec:correspondence_matching}.

\subsection{Implementation details}
%Given our limited compute resources, we cannot afford to experiment with other multi-stage vision transformers, nor can we run large models or run for too many epochs.
Our backbone of choice is the Swin transformer \cite{swin} as it outputs the necessary local-representations required for our local loss.  We follow the implementation details from \cite{dino} and \cite{esvit}. We use the \textit{adamw} optimizer \cite{adamw} with a batch size of 512 and train for a total of 300 epochs. The learning rate is linearly increased during the first 10 epochs to its maximum value of $0.0005*\text{batchsize}/256$ as proposed by \cite{linearlrrule}. It is then reduced throughout the training with a cosine schedule \cite{cosine_schedule}. We also use the sharpening and centering tricks from \cite{dino} to avoid collapse. Regarding the augmentations, we use two global- and 8 local-crops (see appendix). We refer the reader to \cite{dino} for more details.
\begin{figure*}[t]
  \centering
   \includegraphics[width=1\linewidth]{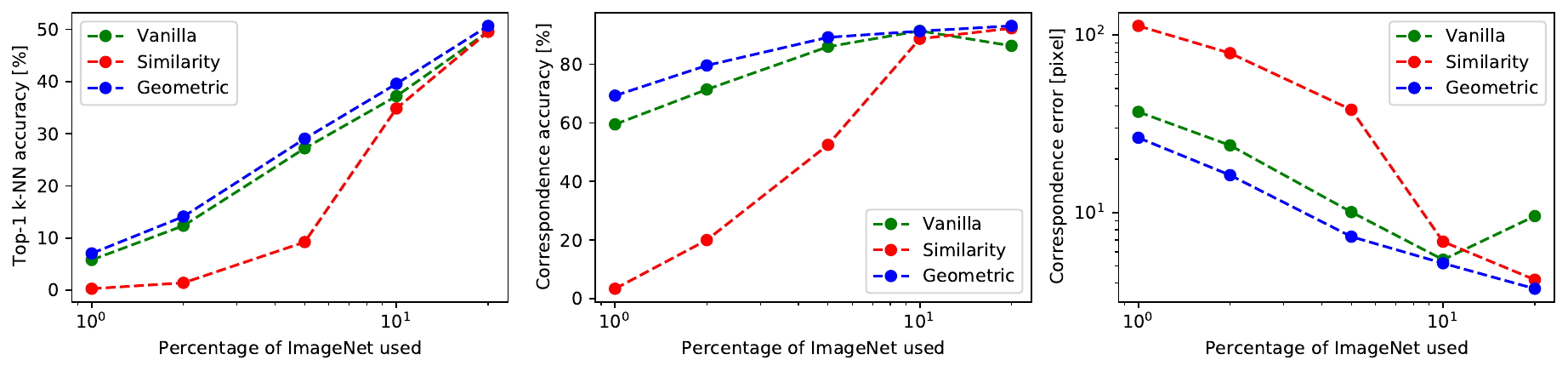}

   \caption{\textbf{Comparison of the performance between the three different settings on ImageNet-1k subsets.} Data points on the x-axis are at 1\%, 2\%, 5\%, 10\%, 20\%. The left plot shows the Top-1 k-NN accuracy. The two other plots are linked to the correspondence matching \Cref{sec:correspondence_matching} where the center and right plot respectively correspond to the accuracy and the error. The \texttt{Geometric} setting shows better performance in all metrics.}
   \label{fig:subsets}
   \vspace{-0.4cm}
\end{figure*}

% \begin{figure}
%     \centering
%     \begin{minipage}{0.48\textwidth}
%         \centering
%         \includegraphics[width=0.999\textwidth]{fig/subsets.pdf} % first figure itself
%         \caption{Pairwise similarity visualization between the large blue augmentation and many smaller augmentations. The 5 most and least similar augmentations are shown respectively in green and red. This corroborates the fact that the learned representation carries the semantics of a dog. Note that this is done fully unsupervised.}
%         \label{fig:densenodense}
%     \end{minipage}\hfill
%     \begin{minipage}{0.48\textwidth}
%         \centering
%         \includegraphics[width=0.999\textwidth]{fig/subsets_error_acc.pdf} % second figure itself
%         \caption{Pairwise similarity visualization between the large blue augmentation and many smaller augmentations. The 5 most and least similar augmentations are shown respectively in green and red. This corroborates the fact that the learned representation carries the semantics of a dog. Note that this is done fully unsupervised.}
%         \label{fig:simsim}
%     \end{minipage}
% \end{figure}

\subsection{Evaluation benchmarks}

We follow the two most common ImageNet \cite{imagenet} unsupervised benchmarks from the literature \cite{dino,instance-level-discrimination,moco,byol} \ie the linear and $k$-NN benchmarks. In both cases, the backbone network and MLP-heads are trained on the training set without using labels. For the linear evaluation, a linear layer is added on top of the frozen global-representation $\glob$ and is trained using the training set (data-augmentations are used) including the labels. The classification accuracy on the test set is evaluated using a center-crop of 224x224. This evaluation protocol is quite computationally intensive as the model needs to compute a forward pass for multiple epochs. The $k$-NN benchmark on the other hand only needs one pass. For each image in both the training and test set, the global-representation of a center-crop (224x224) is computed. Then, each image from the test set gets a label assigned based on a vote from the $k$ nearest neighbors in the training set (anchor points). We use $k=20$ to stay consistent with previous works.%(unless unless stated otherwise)

% \input{fig/distance_temp_ablation}
%In particular, we obtain >+20\% w.r.t. EsViT when training on the 5\% subset. We hypothesize that this is due to \TODO{explain more when more data is here}

\begin{table}[t]
\definecolor{darkgreen}{rgb}{0, 0.5, 0}
\centering
\caption{\textbf{Comparison of multiple methods with similar throughput with ImageNet-1k pretraining.} Rows in blue are results coming from our own runs to study the benefit of the additional local self-distillation.}
\vspace{0.2cm}
\resizebox{\columnwidth}{!}{%
\begin{tabular}{lcccccc}
\hline
\textbf{Method} & \textbf{Backbone}   & \textbf{\#Params} & \textbf{FLOPS} & \textbf{\#Epochs} & \textbf{Linear}      & \textbf{k-NN} \\
\hline
% BYOL  \cite{byol} & ResNet-200 (x2)      & 250M    & G  & 1000     & 79.6         & -            \\
% DINO \cite{dino} & ViT-B/P=8 & 85M    & G    & 800 & 80.1            &    77.4            \\
% EsViT \cite{esvit} & Swin-B/W=14 & 87M    & G    & 300    & \textbf{81.3}            &     \textbf{79.3}         \\

\hline
SimCLR \cite{pmlr-v119-chen20j} & ResNet-50      & 24M  & 4B    & 800    & 69.3              & -           \\
SimCLR v2 \cite{simclrv2}        & ResNet-50      & 24M  & 4B   & 800    & 71.7              & -          \\
BYOL  \cite{byol} & ResNet-50   & 24M   & 4B  & 1000     & 74.3      & -          \\
DINO  \cite{dino} & ViT-S/P=16      & 21M &   4.6B   & 800  & 77.0              & 74.5          \\
MoCo v3 \cite{mocov3} & ViT-S/P=16      & 21M &   4.6B   & 600  & 73.4              & -          \\
EsViT \cite{esvit} & Swin-T/W=7 & 28M  & 4.5B & 300     & 78.0       &     75.7          \\

\hline
\rowcolor{blue!5} \texttt{Vanilla}  & Swin-T/W=7 & 28M & 4.5B  & 300    &     77.0 &  74.2        \\   
\rowcolor{blue!5}  \texttt{Similarity}\footnotemark & Swin-T/W=7 & 28M  & 4.5B & 300     & 77.9    \textcolor{darkgreen}{(+ 0.9)}   &     75.3     \textcolor{darkgreen}{(+ 1.1)}     \\
\rowcolor{blue!5} \texttt{Geometric}  & Swin-T/W=7 & 28M & 4.5B  & 300 & 77.8 \textcolor{darkgreen}{(+ 0.8)}& 75.4 \textcolor{darkgreen}{(+ 1.2)}    
\end{tabular}%
}
\label{tab:main_table}
\end{table}
\footnotetext{In theory, this row should match the the row EsViT \cite{esvit}. However, the authors of \cite{esvit} do not report the downstream evaluations of the last epoch but select the best epoch. This explains the slightly lower performance of the blue row. More details can be found on their \href{https://github.com/microsoft/esvit/commit/c84b67ad25f15d6c706b13f25527009632dca11d}{Github}. All evaluations from our own runs evaluate the model after the final epoch of pretraining.}
\subsection{ImageNet-1k}
\label{sec:imagenet1k}
Both the linear and $k$-NN benchmark results are reported in \Cref{tab:main_table}. The first block of rows compares previous works (including SOTA) with backbones of similar computational requirements. These include ResNet-50 \cite{resnet}, ViT-Small \cite{vit16x16} and Swin-Tiny \cite{swin}. The second block of rows (in blue) are results coming from our own runs to study the benefit of the additional local cues. These runs were trained for 7 days on 8x NVIDIA A100. The \texttt{Similarity} and \texttt{Geometric} matchings outperform the \texttt{Vanilla} method which enforces coherence only at the global level confirming that the additional local regularization is helpful.

\subsection{Other datasets}
\label{sec:otherdatasets}
To get a better idea of the robustness of the additional self-supervised loss at the local level, we train all methods on other datasets. We introduce the local loss to get stronger self-supervision and more efficient learning, which is best studied by looking at the behavior on small scale datasets. These include an artificial setting where we sample 1\%, 2\%, 5\%, 10\% and 20\% subsets of ImageNet-1k in order to evaluate the relative performances in low-data regimes. Even though this is an artificial setting, these subsets are well curated, image-centered and contain a lot of diversity making them ideal for such a study. We also include evaluations on 3 other datasets: \textit{Food-101} \cite{food101}, \textit{NCT-CRC-HE-100K} \cite{nct} and ImageNet-100 (100 class subset of ImageNet-1k).

\begin{table}[b]
\centering
\caption{\textbf{Performance comparison of \texttt{Vanilla}, \texttt{Similarity} and \texttt{Geometric} on the $k$-NN and linear evaluation benchmarks.} Rows in different shades of gray are trained on the training set of Food-101, NCT-CRC-HE-100K, ImageNet-100 and evaluated on the corresponding test set. Each shade of gray represents a different model (backbones from top to bottom: Swin-T/7x7, Swin-T/14x14 and Swin-S/7x7). As a point of reference, the blue rows are trained on ImageNet-1k and evaluated similarly (backbone: Swin-T/7x7). NA entries mean the training crashed due to numerical instabilities.} %test set of Food-101, NCT-CRC-HE-100K, ImageNet-100.}
% \vspace{0.2cm}
\resizebox{\columnwidth}{!}{

\begin{tabular}{l c c c c c c}
\cmidrule(lr){2-7}
& \multicolumn{2}{c}{\texttt{Vanilla}} & \multicolumn{2}{c}{\texttt{Similarity}} & \multicolumn{2}{c}{\texttt{Geometric}}\\
\cmidrule(lr){2-3} \cmidrule(lr){4-5} \cmidrule(lr){6-7}
& $k$-NN & linear & $k$-NN & linear & $k$-NN & linear \\
\hline
\rowcolor{gray!0} Food-101 & 69.3 & 79.4 & 1.7 \textcolor{red}{(-67.6)}      & 3.0 \textcolor{red}{(-76.4)}      & 73.1 \textcolor{darkgreen}{(+3.8)} & 82.2 \textcolor{darkgreen}{(+2.8)} \\
\rowcolor{gray!0} NCT-CRC-HE-100K & 91.9 & 92.1 & 46.0 \textcolor{red}{(-45.9)}      & 53.5 \textcolor{red}{(-38.6)}      & 89.5 \textcolor{red}{(-2.4)}       & 90.2 \textcolor{red}{(-1.9)}       \\
\rowcolor{gray!0} ImageNet-100 & 76.5 & 82.0 & 76.2 \textcolor{red}{(-0.3)}       & 81.4 \textcolor{red}{(-0.6)}       & 79.5 \textcolor{darkgreen}{(+3.0)} & 84.4 \textcolor{darkgreen}{(+2.4)} \\

\hline
\rowcolor{gray!5} Food-101        & 69.5 & 78.5 & 0.9 \textcolor{red}{(-68.6)}  & 2.1 \textcolor{red}{(-76.4)}  & 72.6 \textcolor{darkgreen}{(+3.1)} & 80.7 \textcolor{darkgreen}{(+2.2)} \\
\rowcolor{gray!5} NCT-CRC-HE-100K & 90.8 & 90.1 & 44.1 \textcolor{red}{(-46.7)} & 34.1 \textcolor{red}{(-56.0)} & 91.0 \textcolor{darkgreen}{(+0.2)}   & 89.4 \textcolor{red}{(-0.7)} \\
\rowcolor{gray!5} ImageNet-100    & 76.8 & 81.6 & 2.1 \textcolor{red}{(-74.7)}  & 1.9 \textcolor{red}{(-79.7)}  & 78.9 \textcolor{darkgreen}{(+2.1)} & 83.0 \textcolor{darkgreen}{(+1.4)} \\

\hline

\rowcolor{gray!15} Food-101        & 70.9 & 79.6 & NA & NA    & 73.8 \textcolor{darkgreen}{(+2.9)} & 82.5 \textcolor{darkgreen}{(+2.9)} \\
\rowcolor{gray!15} NCT-CRC-HE-100K & 90.8 & 89.4 & NA  & NA   & 90.5 \textcolor{red}{(-0.3)} & 86.9 \textcolor{red}{(-2.5)} \\
\rowcolor{gray!15} ImageNet-100    & 78.6 & 83.1 & 76.1 \textcolor{red}{(-2.5)} & 80.9 \textcolor{red}{(-2.2)} & 80.2 \textcolor{darkgreen}{(+1.6)} & 84.1 \textcolor{darkgreen}{(+1.0)} \\

\hline
\rowcolor{blue!5} Food-101 & 67.7 & 81.4 & 68.4 \textcolor{darkgreen}{(+0.7)} & 82.2 \textcolor{darkgreen}{(+0.8)} & 68.6 \textcolor{darkgreen}{(+0.9)} & 82.2 \textcolor{darkgreen}{(+0.8)} \\
\rowcolor{blue!5} NCT-CRC-HE-100K & 89.2 & 93.3 & 90.8 \textcolor{darkgreen}{(+1.6)} & 93.1 \textcolor{red}{(-0.2)}       & 90.3 \textcolor{darkgreen}{(+1.1)} & 94.2 \textcolor{darkgreen}{(+0.9)} \\
\rowcolor{blue!5} ImageNet-100 & 86.2 & 88.1 & 87.0 \textcolor{darkgreen}{(+0.8)} & 88.8 \textcolor{darkgreen}{(+0.7)} & 87.5 \textcolor{darkgreen}{(+1.3)} & 88.3 \textcolor{darkgreen}{(+0.2)}

\end{tabular}
}
\label{table:trained_on_multiple}
\end{table}

\subsubsection{ImageNet-1k subsampled}
The 1\%, 2\%, 5\%, 10\% and 20\% subsets are obtained by sampling respectively 10, 20, 50, 100, 200 images from each class to avoid imbalances. Each method is independently trained on a subset and evaluated on the $k$-NN benchmark. Note that this evaluation can be done using the full training set or the training subsets. Since we are using very little training data (\eg 1\%), we choose to evaluate the anchor points on the full training data to make the metric more robust and fair across all subsets. There are two main observations from the left plot of \Cref{fig:subsets}: 1) incorporating local cues using similarity matchings is hurtful for small subsets and 2) geometric matchings on the other hand are robust and provide additional accuracy on all subsets. Similarities between local-representations in low-data regimes are mostly based on low-level features leading to collapse of the matching function. We will confirm this in the following section. Note that the y-scale of the left plot of \Cref{fig:subsets} goes from 0 to 50\%: the relative difference between the vanilla and the geometric matching method is in order of 1.5\% which is highly significant on this benchmark. Still, the size of the dataset remains the dominant factor and  regularization based on local correspondences cannot replace that.

\subsubsection{Smaller scale datasets \& different models}
The evaluation on \textit{Food-101} \cite{food101}, \textit{NCT-CRC-HE-100K} \cite{nct} and ImageNet-100 with different models (Swin-T/7x7, Swin-T/14x14 and Swin-S/7x7) can be found in \Cref{table:trained_on_multiple}. The datasets are chosen because they all contain about 100k images and images have a resolution similar to ImageNet-1k. Rows in light gray are trained from scratch on the training set of Food-101, NCT-CRC-HE-100K, ImageNet-100 and evaluated on the corresponding test set. Rows in darker gray are trained on ImageNet-1k and evaluated on the test set of Food-101, NCT-CRC-HE-100K, ImageNet-100. In most evaluations, the additional geometric local regularization improves the performance on the downstream tasks. A key finding from \Cref{table:trained_on_multiple} is that the similarity based local loss used in SOTA work EsViT \cite{esvit} is hurtful in low-data regimes while the geometric local loss is robust and shows improvements in almost all evaluations. The takeaways are similar irrespective of model size and window size. The performance drop due to collapse (see \cref{sec:collapse}) does a appear a bit worse when using Swin-T/14x14 compared to Swin-T/7x7. This can be explained by the fact that a larger window size allows tokens to attend to a wider set of tokens in general, or in particular, a wider set of similar tokens during the transition to a collapsed state.

\begin{figure}[t]
  \centering
   \includegraphics[width=\linewidth]{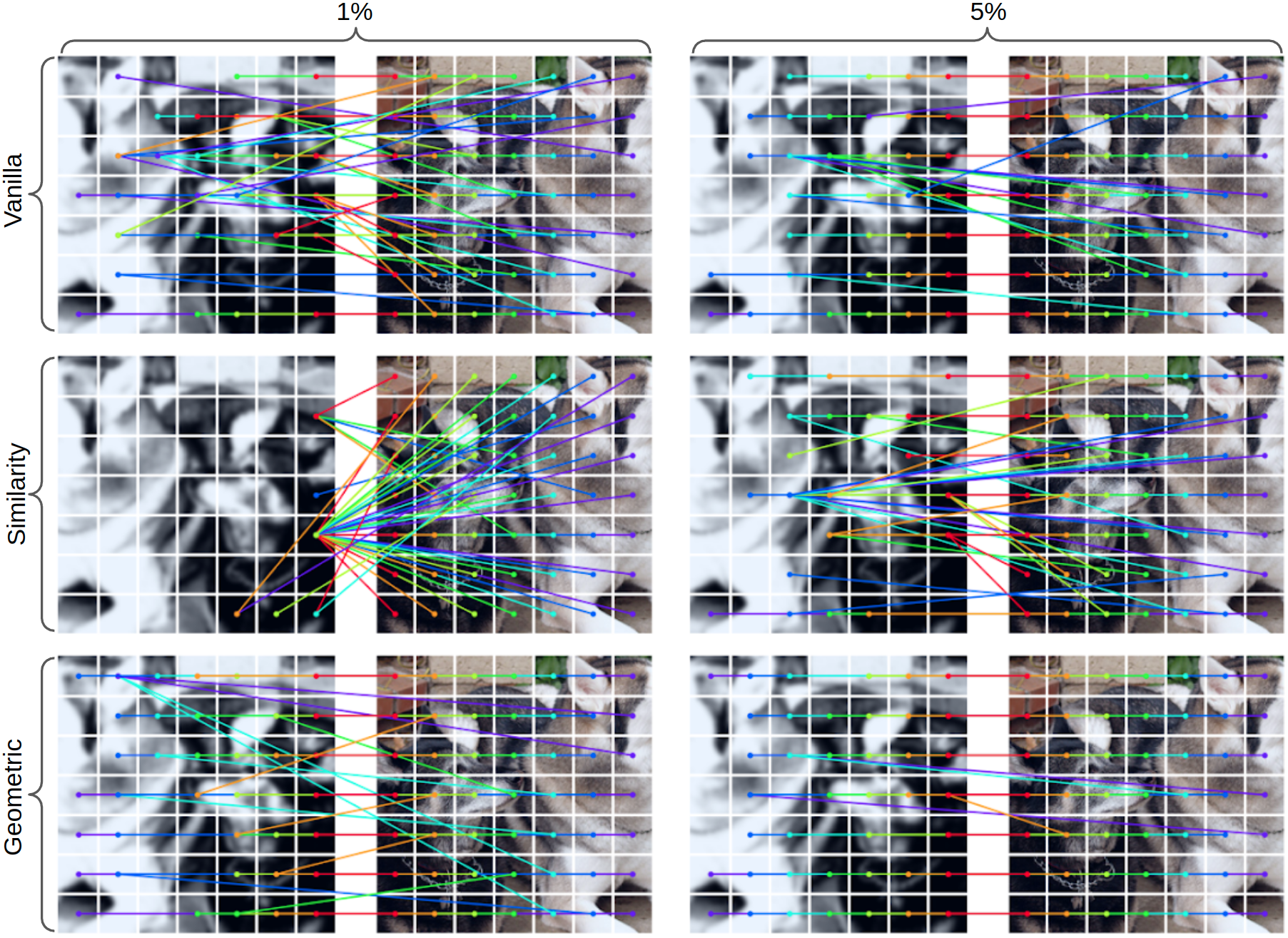}

   \caption{\textbf{Visual comparison of the similarity correspondence between two augmentations.} This is done for \texttt{Vanilla}, \texttt{Similarity} and \texttt{Geometric} settings on 2 training data regimes: 1\% and 5\% of the full ImageNet training data. Every location on the right side is matched to a location on the left image based on the distance in the learned feature space. Colors are used only to better distinguish different matchings. Best viewed in color and zoomed in.}
   \label{fig:subset_matching}
%   \vspace{-0.4cm}
\end{figure}

% \begin{figure*}[t!]
%   \centering
%   %\fbox{\rule{0pt}{2in} \rule{0.9\linewidth}{0pt}}
%   \includegraphics[width=0.8\linewidth]{fig/cvpr.drawio.png}

%   \caption{\textbf{Example of our inverse distance-based weighted local-representation matching between 2 augmentations of an input image.} Each location on a coarse grained grid (corresponding to the area of an output token) on the right image is linked to the best matching location on the left image based on the distance between the center of the locations. The size of the matching line corresponds to the weighting which is computed relatively to all others matching distances. The smaller the matching distance is in the input image, the higher the weighting is. Colors are used only to better differentiate different matchings.}
%   \label{fig:first_fig}
% \end{figure*}

% \input{fig/collapse_esvit}

\subsection{Collapse of the similarity matching function}
\label{sec:collapse}
In general, self-supervised methods (even with a single global loss) are prone to mode collapse since they cannot use labels as targets for their outputs and instead have to bootstrap their own outputs during training. As such, a lot of care has to be put in the design of the training algorithm. In contrastive learning methods, a loss function with appropriate negative samples mitigates the issue \cite{pmlr-v119-chen20j}. Analogously in self-distillation methods, a careful tuning of the temperature parameters in the centering and sharpening trick is required \cite{dino}. In a dual global-local objective framework, collapse can also occur at the local level. Collapse at the local level can happen when using the \texttt{Similarity} setting proposed in \cite{esvit}. Such failure cases are shown in the appendix. Due to the nature of the backbone, collapse at the local level implies collapse at the global level. That is because the global-representation $\glob = \bar{f}(\im)$ is a direct function of the dense representation $\dense$ i.e. $\glob = g(\dense)$ with $g$ an average pooling layer or attention layer. When collapse occurs (both global and local), we get that $\nabla_{\im} \bar{f}(\im) \approx \boldsymbol{0}$. That is, the model discards all information from the input image $\im$ leading to downstream evaluations close to a random accuracy of $\frac{1}{nb\_class}$ as can be seen in some entries of the \texttt{Similarity} column of \Cref{table:trained_on_multiple}.

% this is the case for crop_i <-> crop_j

% empirically, we show that it also holds for any pair of images from the dataset i.e. all (augmented) images map to the same representation (global and local)
% sum of distance squared of all

% when that is the case, we mathematically get that 
% grad_x f(x) ~= 0 (model discards all information from input, downstream accuracy is almost random)
% this can be observed in Table where some entries are very close to random accuracy (1/nb_class)

If collapse of the matching function occurs, the method cannot recover because $\mathcal{L}_L^{sim}$ is enforced making the collapse even worse. The \texttt{Geometric} setting avoids this issue by construction leading to a more robust training.

\subsection{Correspondence matching based on similarity}
\label{sec:correspondence_matching}
We analyze the learned representation by looking at the quality of correspondence matching based on the similarities of local-representations. Two augmentations of an input image from the validation set are computed using a tweaked data augmentation pipeline where there always exists an exact correspondence mapping between the local-representations \ie augmentations are always cropped and resized in the same manner. This spatial correspondence mapping is used as ground truth and we evaluate the matchings obtained using token similarity for all three settings. In the center and right part of Figure \ref{fig:comparison_matchings}, the results are shown using two metrics: 1) the classification accuracy (i.e., how many of the local representations are matched correctly) and 2) the distance error, both w.r.t. the ground truth correspondence mapping. The geometric matchings processes images in a way that better preserves the spatial information. Qualitative evaluations can be found in \Cref{fig:subset_matching}.

%it happens in same case that one local-representation is matched to many local-representation from the other crop.

%Since the quantity of data is quite significant, we run the $k$-NN evaluation only using the training subset of 10\%. 
\section{Conclusion \& Future work}
% We propose a new method titled \textbf{SPADE} (\textbf{SP}atial-\textbf{A}ware and \textbf{D}ata-\textbf{E}fficient) which uses the spatial information from the original input images to enforce spatial consistency between the local-representations of a multi-stage vision transformer. Our method shows advantages over previous work both in terms of 1) higher performance on downstream tasks and 2) better preservation of the spatial relations of the input images. This is particularly true in low-data regimes. We hope this work can encourage other researchers working with custom small-scale datasets to try our method and confirm the advantages of our method in the low-data regimes using subsets of ImageNet, as well as inspire other researchers with more compute power to upscale our method on larger backbones.

Self-supervised training of visual transformers using self-distillation is becoming the standard way of obtaining visual representation from images by solving a proxy task at the image-level (global-level). We can leverage additional self-supervision by incorporating self-distillation at the local-level. This is done by enforcing coherence between pairs of local-representations (acts as a regularizer). We observe an improvement on downstream tasks using multiple datasets. We study the effect of the matching function used to generate pairs of local-representations from both augmentations. A geometry based matching function shows advantages over a similarity based matching function both in terms of 1) higher performance on downstream tasks and 2) better preservation of the spatial relations of the input images. This is particularly true in low-data regimes, in which case we observe a collapse of the similarity based matching function in some settings. We believe the insights from this paper can lead to a better crafting of a data-driven local-representation matching function to explicitly avoid collapse and that an upscaling of these methods to very large backbones can surpass the current state of the art.

% We hope this work can incentivize the research community to incorporate additional self-supervised losses at the local level to obtain richer gradients leading to better performance on the downstream tasks.

% \begin{figure}[h]
%   \centering
%   %\fbox{\rule{0pt}{2in} \rule{0.9\linewidth}{0pt}}
%   \includegraphics[width=\linewidth]{fig/similarity_food101.png}

%   \caption{}
%   \label{fig:nolabel}
% \end{figure}

\section{Acknowledgements}
This paper is part of a project that has received funding from the European Research Council (ERC) under the European Union’s Horizon 2020 research and innovation programme (Grant agreement No. 101021347). The computational resources and services used in this work were provided by the VSC (Flemish Supercomputer Center), funded by the Research Foundation Flanders (FWO) and the Flemish Government – department EWI.

%%%%%%%%% BODY TEXT
{\small
\bibliographystyle{ieee_fullname}
\bibliography{egbib}

\begin{thebibliography}{10}\itemsep=-1pt

\bibitem{NEURIPS2019_ddf35421}
Philip Bachman, R~Devon Hjelm, and William Buchwalter.
\newblock Learning representations by maximizing mutual information across
  views.
\newblock In H. Wallach, H. Larochelle, A. Beygelzimer, F. d\textquotesingle
  Alch\'{e}-Buc, E. Fox, and R. Garnett, editors, {\em Advances in Neural
  Information Processing Systems}, volume~32. Curran Associates, Inc., 2019.

\bibitem{masked_image_modeling}
Hangbo Bao, Li Dong, and Furu Wei.
\newblock Beit: {BERT} pre-training of image transformers.
\newblock {\em CoRR}, abs/2106.08254, 2021.

\bibitem{predicting_noise}
Piotr Bojanowski and Armand Joulin.
\newblock Unsupervised learning by predicting noise.
\newblock In {\em Proceedings of the 34th International Conference on Machine
  Learning}, volume~70, pages 517--526. PMLR, 06--11 Aug 2017.

\bibitem{food101}
Lukas Bossard, Matthieu Guillaumin, and Luc Van~Gool.
\newblock Food-101 -- mining discriminative components with random forests.
\newblock In David Fleet, Tomas Pajdla, Bernt Schiele, and Tinne Tuytelaars,
  editors, {\em Computer Vision -- ECCV 2014}, pages 446--461, Cham, 2014.
  Springer International Publishing.

\bibitem{swav}
Mathilde Caron, Ishan Misra, Julien Mairal, Priya Goyal, Piotr Bojanowski, and
  Armand Joulin.
\newblock Unsupervised learning of visual features by contrasting cluster
  assignments.
\newblock {\em CoRR}, abs/2006.09882, 2020.

\bibitem{dino}
Mathilde Caron, Hugo Touvron, Ishan Misra, Hervé Jégou, Julien Mairal, Piotr
  Bojanowski, and Armand Joulin.
\newblock Emerging properties in self-supervised vision transformers.
\newblock {\em CoRR}, abs/2104.14294, 2021.

\bibitem{pmlr-v119-chen20j}
Ting Chen, Simon Kornblith, Mohammad Norouzi, and Geoffrey Hinton.
\newblock A simple framework for contrastive learning of visual
  representations.
\newblock In Hal~Daumé III and Aarti Singh, editors, {\em Proceedings of the
  37th International Conference on Machine Learning}, volume 119 of {\em
  Proceedings of Machine Learning Research}, pages 1597--1607. PMLR, 13--18 Jul
  2020.

\bibitem{simclrv2}
Ting Chen, Simon Kornblith, Kevin Swersky, Mohammad Norouzi, and Geoffrey~E.
  Hinton.
\newblock Big self-supervised models are strong semi-supervised learners.
\newblock {\em CoRR}, abs/2006.10029, 2020.

\bibitem{mocov2}
Xinlei Chen, Haoqi Fan, Ross Girshick, and Kaiming He.
\newblock Improved baselines with momentum contrastive learning.
\newblock {\em arXiv preprint arXiv:2003.04297}, 2020.

\bibitem{simsiam}
Xinlei Chen and Kaiming He.
\newblock Exploring simple siamese representation learning.
\newblock {\em CoRR}, abs/2011.10566, 2020.

\bibitem{mocov3}
Xinlei Chen, Saining Xie, and Kaiming He.
\newblock An empirical study of training self-supervised vision transformers.
\newblock {\em CoRR}, abs/2104.02057, 2021.

\bibitem{conditional_posenc}
Xiangxiang Chu, Bo Zhang, Zhi Tian, Xiaolin Wei, and Huaxia Xia.
\newblock Do we really need explicit position encodings for vision
  transformers?
\newblock {\em CoRR}, abs/2102.10882, 2021.

\bibitem{ciga2021resource}
Ozan Ciga, Tony Xu, and Anne~L. Martel.
\newblock Resource and data efficient self supervised learning, 2021.

\bibitem{imagenet}
J. Deng, W. Dong, R. Socher, L.-J. Li, K. Li, and L. Fei-Fei.
\newblock {ImageNet: A Large-Scale Hierarchical Image Database}.
\newblock In {\em CVPR09}, 2009.

\bibitem{DBLP:journals/corr/abs-2103-04814}
Jian Ding, Enze Xie, Hang Xu, Chenhan Jiang, Zhenguo Li, Ping Luo, and
  Gui{-}Song Xia.
\newblock Unsupervised pretraining for object detection by patch
  reidentification.
\newblock {\em CoRR}, abs/2103.04814, 2021.

\bibitem{contex_pred_doersch}
C. {Doersch}, A. {Gupta}, and A.~A. {Efros}.
\newblock Unsupervised visual representation learning by context prediction.
\newblock In {\em 2015 IEEE International Conference on Computer Vision
  (ICCV)}, pages 1422--1430, 2015.

\bibitem{vit16x16}
Alexey Dosovitskiy, Lucas Beyer, Alexander Kolesnikov, Dirk Weissenborn,
  Xiaohua Zhai, Thomas Unterthiner, Mostafa Dehghani, Matthias Minderer, Georg
  Heigold, Sylvain Gelly, Jakob Uszkoreit, and Neil Houlsby.
\newblock An image is worth 16x16 words: Transformers for image recognition at
  scale.
\newblock {\em CoRR}, abs/2010.11929, 2020.

\bibitem{gidaris_rotations}
Spyros Gidaris, Praveer Singh, and Nikos Komodakis.
\newblock Unsupervised representation learning by predicting image rotations.
\newblock {\em CoRR}, abs/1803.07728, 2018.

\bibitem{linearlrrule}
Priya Goyal, Piotr Doll{\'{a}}r, Ross~B. Girshick, Pieter Noordhuis, Lukasz
  Wesolowski, Aapo Kyrola, Andrew Tulloch, Yangqing Jia, and Kaiming He.
\newblock Accurate, large minibatch {SGD:} training imagenet in 1 hour.
\newblock {\em CoRR}, abs/1706.02677, 2017.

\bibitem{byol}
Jean-Bastien Grill, Florian Strub, Florent Altché, Corentin Tallec, Pierre~H.
  Richemond, Elena Buchatskaya, Carl Doersch, Bernardo~Avila Pires,
  Zhaohan~Daniel Guo, Mohammad~Gheshlaghi Azar, Bilal Piot, Koray Kavukcuoglu,
  Rémi Munos, and Michal Valko.
\newblock Bootstrap your own latent: A new approach to self-supervised
  learning, 2020.

\bibitem{mae}
Kaiming He, Xinlei Chen, Saining Xie, Yanghao Li, Piotr Doll{\'{a}}r, and
  Ross~B. Girshick.
\newblock Masked autoencoders are scalable vision learners.
\newblock {\em CoRR}, abs/2111.06377, 2021.

\bibitem{he2021masked}
Kaiming He, Xinlei Chen, Saining Xie, Yanghao Li, Piotr Dollár, and Ross
  Girshick.
\newblock Masked autoencoders are scalable vision learners, 2021.

\bibitem{moco}
Kaiming He, Haoqi Fan, Yuxin Wu, Saining Xie, and Ross~B. Girshick.
\newblock Momentum contrast for unsupervised visual representation learning.
\newblock {\em CoRR}, abs/1911.05722, 2019.

\bibitem{maskrcnn}
Kaiming He, Georgia Gkioxari, Piotr Doll{\'{a}}r, and Ross~B. Girshick.
\newblock Mask {R-CNN}.
\newblock {\em CoRR}, abs/1703.06870, 2017.

\bibitem{resnet}
Kaiming He, Xiangyu Zhang, Shaoqing Ren, and Jian Sun.
\newblock Deep residual learning for image recognition.
\newblock In {\em Proceedings of the IEEE Conference on Computer Vision and
  Pattern Recognition (CVPR)}, June 2016.

\bibitem{hjelm2019learning}
Devon Hjelm, Alex Fedorov, Samuel Lavoie-Marchildon, Karan Grewal, Philip
  Bachman, Adam Trischler, and Yoshua Bengio.
\newblock Learning deep representations by mutual information estimation and
  maximization.
\newblock In {\em ICLR 2019}. ICLR, April 2019.

\bibitem{nct}
Jakob Kather, Johannes Krisam, Pornpimol Charoentong, Tom Luedde, Esther
  Herpel, Cleo-Aron Weis, Timo Gaiser, Alexander Marx, Nek Valous, Dyke Ferber,
  Lina Jansen, Constantino Reyes-Aldasoro, Inka Zoernig, Dirk Jäger, Hermann
  Brenner, Jenny Chang-Claude, Michael Hoffmeister, and Niels Halama.
\newblock Predicting survival from colorectal cancer histology slides using
  deep learning: A retrospective multicenter study.
\newblock {\em PLoS Medicine}, 16:e1002730, 01 2019.

\bibitem{esvit}
Chunyuan Li, Jianwei Yang, Pengchuan Zhang, Mei Gao, Bin Xiao, Xiyang Dai, Lu
  Yuan, and Jianfeng Gao.
\newblock Efficient self-supervised vision transformers for representation
  learning.
\newblock {\em CoRR}, abs/2106.09785, 2021.

\bibitem{https://doi.org/10.48550/arxiv.2111.11429}
Y. Li, S. Xie, X. Chen, P. Dollar, K. He, and R. Girshick.
\newblock Benchmarking detection transfer learning with vision transformers,
  2021.

\bibitem{liu2021efficient}
Yahui Liu, Enver Sangineto, Wei Bi, Nicu Sebe, Bruno Lepri, and Marco~De Nadai.
\newblock Efficient training of visual transformers with small-size datasets,
  2021.

\bibitem{swin}
Ze Liu, Yutong Lin, Yue Cao, Han Hu, Yixuan Wei, Zheng Zhang, Stephen Lin, and
  Baining Guo.
\newblock Swin transformer: Hierarchical vision transformer using shifted
  windows.
\newblock {\em CoRR}, abs/2103.14030, 2021.

\bibitem{cosine_schedule}
Ilya Loshchilov and Frank Hutter.
\newblock {SGDR:} stochastic gradient descent with restarts.
\newblock {\em CoRR}, abs/1608.03983, 2016.

\bibitem{adamw}
Ilya Loshchilov and Frank Hutter.
\newblock Fixing weight decay regularization in adam, 2018.

\bibitem{contex_pred_Mundhenk}
T.~N. {Mundhenk}, D. {Ho}, and B.~Y. {Chen}.
\newblock Improvements to context based self-supervised learning.
\newblock In {\em 2018 IEEE/CVF Conference on Computer Vision and Pattern
  Recognition}, pages 9339--9348, 2018.

\bibitem{noroozi_jigsaw}
Mehdi Noroozi and Paolo Favaro.
\newblock Unsupervised learning of visual representations by solving jigsaw
  puzzles.
\newblock In Bastian Leibe, Jiri Matas, Nicu Sebe, and Max Welling, editors,
  {\em Computer Vision -- ECCV 2016}, pages 69--84, Cham, 2016. Springer
  International Publishing.

\bibitem{inpainting}
Deepak Pathak, Philipp Kr{\"{a}}henb{\"{u}}hl, Jeff Donahue, Trevor Darrell,
  and Alexei~A. Efros.
\newblock Context encoders: Feature learning by inpainting.
\newblock {\em CoRR}, abs/1604.07379, 2016.

\bibitem{unsupervised_learning_of_dense_visual_rep}
Pedro~O. Pinheiro, Amjad Almahairi, Ryan~Y. Benmalek, Florian Golemo, and
  Aaron~C. Courville.
\newblock Unsupervised learning of dense visual representations.
\newblock In {\em NeurIPS}, 2020.

\bibitem{ranftl2021vision}
René Ranftl, Alexey Bochkovskiy, and Vladlen Koltun.
\newblock Vision transformers for dense prediction, 2021.

\bibitem{saillard2021self}
Charlie Saillard, Olivier Dehaene, Tanguy Marchand, Olivier Moindrot, Aurélie
  Kamoun, Benoit Schmauch, and Simon Jegou.
\newblock Self supervised learning improves dmmr/msi detection from histology
  slides across multiple cancers, 2021.

\bibitem{relative_positional_encoding}
Peter Shaw, Jakob Uszkoreit, and Ashish Vaswani.
\newblock Self-attention with relative position representations.
\newblock {\em CoRR}, abs/1803.02155, 2018.

\bibitem{whatmakesforgoodviews}
Yonglong Tian, Chen Sun, Ben Poole, Dilip Krishnan, Cordelia Schmid, and
  Phillip Isola.
\newblock What makes for good views for contrastive learning.
\newblock {\em CoRR}, abs/2005.10243, 2020.

\bibitem{all_you_need_is_attention}
Ashish Vaswani, Noam Shazeer, Niki Parmar, Jakob Uszkoreit, Llion Jones,
  Aidan~N Gomez, \L~ukasz Kaiser, and Illia Polosukhin.
\newblock Attention is all you need.
\newblock In I. Guyon, U.~V. Luxburg, S. Bengio, H. Wallach, R. Fergus, S.
  Vishwanathan, and R. Garnett, editors, {\em Advances in Neural Information
  Processing Systems}, volume~30, pages 5998--6008. Curran Associates, Inc.,
  2017.

\bibitem{Wang_2021_CVPR}
Xinlong Wang, Rufeng Zhang, Chunhua Shen, Tao Kong, and Lei Li.
\newblock Dense contrastive learning for self-supervised visual pre-training.
\newblock In {\em Proceedings of the IEEE/CVF Conference on Computer Vision and
  Pattern Recognition (CVPR)}, pages 3024--3033, June 2021.

\bibitem{cvt}
Haiping Wu, Bin Xiao, Noel Codella, Mengchen Liu, Xiyang Dai, Lu Yuan, and Lei
  Zhang.
\newblock Cvt: Introducing convolutions to vision transformers, 2021.

\bibitem{instance-level-discrimination}
Zhirong Wu, Yuanjun Xiong, Stella~X. Yu, and Dahua Lin.
\newblock Unsupervised feature learning via non-parametric instance-level
  discrimination.
\newblock {\em CoRR}, abs/1805.01978, 2018.

\bibitem{moby}
Zhenda Xie, Yutong Lin, Zhuliang Yao, Zheng Zhang, Qi Dai, Yue Cao, and Han Hu.
\newblock Self-supervised learning with swin transformers.
\newblock {\em CoRR}, abs/2105.04553, 2021.

\bibitem{pixpro}
Zhenda Xie, Yutong Lin, Zheng Zhang, Yue Cao, Stephen Lin, and Han Hu.
\newblock Propagate yourself: Exploring pixel-level consistency for
  unsupervised visual representation learning.
\newblock In {\em Proceedings of the IEEE/CVF Conference on Computer Vision and
  Pattern Recognition (CVPR)}, pages 16684--16693, June 2021.

\bibitem{DBLP:journals/corr/abs-2111-12309}
Yufei Xu, Qiming Zhang, Jing Zhang, and Dacheng Tao.
\newblock Regioncl: Can simple region swapping contribute to contrastive
  learning?
\newblock {\em CoRR}, abs/2111.12309, 2021.

\bibitem{DBLP:journals/corr/abs-2102-08318}
Ceyuan Yang, Zhirong Wu, Bolei Zhou, and Stephen Lin.
\newblock Instance localization for self-supervised detection pretraining.
\newblock {\em CoRR}, abs/2102.08318, 2021.

\bibitem{focal_transformer}
Jianwei Yang, Chunyuan Li, Pengchuan Zhang, Xiyang Dai, Bin Xiao, Lu Yuan, and
  Jianfeng Gao.
\newblock Focal self-attention for local-global interactions in vision
  transformers.
\newblock {\em CoRR}, abs/2107.00641, 2021.

\bibitem{vil}
Pengchuan Zhang, Xiyang Dai, Jianwei Yang, Bin Xiao, Lu Yuan, Lei Zhang, and
  Jianfeng Gao.
\newblock Multi-scale vision longformer: A new vision transformer for
  high-resolution image encoding, 2021.

\end{thebibliography}
}

\newpage
\appendix
\section*{Appendix}
In a first step, we review the multi-crop strategy \cite{swav}. The pseudo-code for the global-crop algorithm and the multi-crop algorithm can be respectively found in \cref{alg:savit} and \cref{alg:spade-multi}. We then show visualizations of the matchings which are enforced at training time, visualizations of the collapse of the similarity matching function and visualizations of the correspondence mapping between two different images based on the similarity of the local-representations.

\begin{algorithm}[b]
	\caption{Dual global-local self-distillation framework} 
    \textbf{Input:} $\mathcal{X}$: an unlabeled dataset, $N$: the number of augmentations per input image, $\mathbf{P}$: a photometric-augmentation function, $\mathbf{G}$: a geometric-augmentation function, $\mathcal{D}_{aug}$: an augmentation-vector distribution, $f_s$: a backbone student, $f_t$: a backbone teacher, \texttt{OPTIMIZER}: an optimizer\\
    \textbf{Output:} Trained weights $\weight_t$
	\begin{algorithmic}[1]
	\State $\weight_s=\weight_t=\weight_{init}$
	\For{epoch = 1 $\cdots$ NB\_EPOCHS}
		\For {$\im \in \mathcal{X}$}
    		\For {$n \in [N]$}
    		    \State Sample $\augvec^n = [\augvec_{geo}^n, \augvec_{pho}^n]$ from $\mathcal{D}_{aug}$
    		    \State $\aug^n = \mathbf{P}\left(\mathbf{G}(\im, \augvec^n_{geo}), \augvec^n_{pho}\right)$
    		\EndFor
		    \State $\mathcal{W}_{geo} = \{\augvec^1_{geo}, \augvec^2_{geo}, \augvec^3_{geo}, \cdots\}$
		    \State Infer $\mathcal{E}=\{\pos^1, \pos^2, \pos^3, \cdots\}$ from $\mathcal{W}_{geo}$
		    \State $\mathcal{V} = \{\aug^1, \aug^2, \aug^3, \cdots\}$
		    \State $\bar{\mathcal{Z}}_s = \{\bar{f_s}(\aug) : \aug \in \mathcal{V}\}$
		    \State $\bar{\mathcal{Z}}_t = \{\bar{f_t}(\aug) : \aug \in \mathcal{V}\}$
		    \State ${\mathcal{Z}}_s = \{{f_s}(\aug) : \aug \in \mathcal{V}\}$
		    \State ${\mathcal{Z}}_t = \{{f_t}(\aug) : \aug \in \mathcal{V}\}$
		    \State $\mathcal{L}=\mathcal{L}_G+\mathcal{L}_L^{sim/geo}$ \hfill (\cref{eq:global_loss}, \cref{eq:total_geometric_loss}/\cref{eq:total_similarity_loss})
		    \State $\weight_s \leftarrow$ \texttt{OPTIMIZER}$(\weight_s, \nabla_{\weight_s}\mathcal{L})$
		\EndFor
		\State $\weight_t \leftarrow \lambda \weight_t + (1-\lambda) \weight_s$
	\EndFor
	\State \Return $\weight_t$
	\end{algorithmic}
	\label{alg:savit}
\end{algorithm}

\section{Details on Multi-Crop}
Details on the multi-crop strategy \cite{swav} are left out in the main paper for simplicity. For completeness, we give a review of the multi-crop strategy and explicitly explain how it affects our loss terms.

\begin{algorithm}[b]
	\caption{Algorithm 1 edit for multi-crop}
	\begin{algorithmic}[1]
	\setcounter{ALG@line}{3}
	        \State Sample $\augvec^1 = [\augvec_{geo}^1, \augvec_{pho}^1]$ from $\mathcal{D}_{aug_1}^G$
	        \State $\aug^1 = \mathbf{P}\left(\mathbf{S}(\im, \augvec^1_{geo}), \augvec^1_{pho}\right)$
	        \State Sample $\augvec^2 = [\augvec_{geo}^2, \augvec_{pho}^2]$ from $\mathcal{D}_{aug_2}^G$
	        \State $\aug^2 = \mathbf{P}\left(\mathbf{S}(\im, \augvec^2_{geo}), \augvec^2_{pho}\right)$
    		\For {$n \in \{3, 4, \cdots, 2+N_L\}$}
    		    \State Sample $\augvec^n = [\augvec_{geo}^n, \augvec_{pho}^n]$ from $\mathcal{D}_{aug}^L$
    		    \State $\aug^n = \mathbf{P}\left(\mathbf{S}(\im, \augvec^n_{geo}), \augvec^n_{pho}\right)$
    		\EndFor
	\end{algorithmic}
	\label{alg:spade-multi}
\end{algorithm}

\subsection{Review of Multi-Crop}
\begin{figure}[t]
  \centering
   \includegraphics[width=0.7\linewidth]{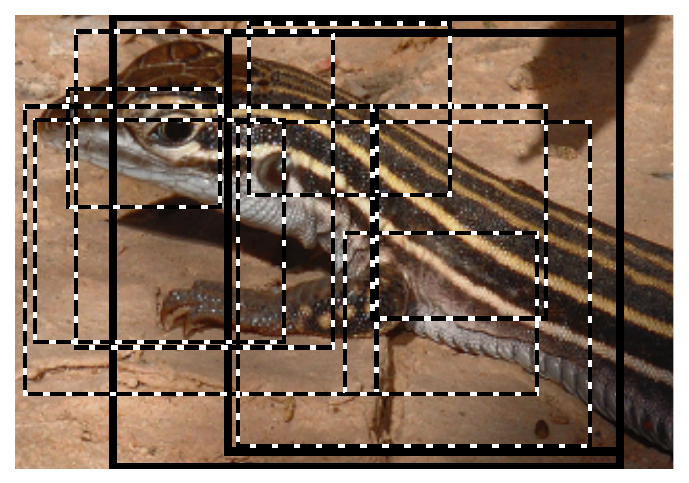}

   \caption{\textbf{Example of the sampled locations using the multi-crop strategy.} 2 global- and 8 local-crops are respectively shown in thick full lines and thin dashed.}
   \label{fig:multi-crop}
\end{figure}

In \Cref{sec:augmentation_pipeline} we explain the data-augmentation pipeline and how multiple augmentations of a single input image are generated. To obtain an augmented image $\aug$ from an input image $\im$, we sample an augmentation vector $\augvec=[\augvec_{geo},\augvec_{pho}]$ from some distribution $\mathcal{D}_{aug}$. This augmentation vector parametrizes both the geometric- and photometric transforms that are applied to $\im$. Spatial transforms include \texttt{CROP}, \texttt{RESIZE} and \texttt{HORIZONTAL\_FLIP} while photometric transforms include \texttt{COLOR\_JITTER}, \texttt{SOLARIZE}, \texttt{GAUSSIAN\_BLUR} and \texttt{GRAYSCALE}. We denote the composition of all geometric transforms by $\mathbf{G}$ and the composition of all photometric transforms by $\mathbf{P}$. In \Cref{sec:augmentation_pipeline}, we assumed that all augmentation vectors $\augvec$ are sampled from the same distribution $\mathcal{D}_{aug}$. The multi-crop strategy \cite{swav} removes this assumption. Instead, we segregate augmentations into two categories, global- and local-crops. Local-crops are taken from smaller regions of the input image while global-crops are taken from larger ones. Local-crops are also resized to $96\times96$ pixels while global-crops are resized to $224\times224$ pixels. This is illustrated in \Cref{fig:multi-crop}.

% \begin{figure*}[t!]
%   \centering
%   %\fbox{\rule{0pt}{2in} \rule{0.9\linewidth}{0pt}}
%   \includegraphics[width=0.8\linewidth]{fig/cvpr.drawio.png}

%   \caption{\textbf{Example of our inverse distance-based weighted local-representation matching between 2 augmentations of an input image.} Each location on a coarse grained grid (corresponding to the area of an output token) on the right image is linked to the best matching location on the left image based on the distance between the center of the locations. The size of the matching line corresponds to the weighting which is computed relatively to all others matching distances. The smaller the matching distance is in the input image, the higher the weighting is. Colors are used only to better differentiate different matchings.}
%   \label{fig:first_fig}
% \end{figure*}

For every single original input image $\im$, 2 global-crops and $N_L=8$ local-crops are generated. All augmentation vectors for local-crops are sampled from the same distribution $\mathcal{D}_{aug}^{L}$ while the augmentation vectors for global-crops are individually sampled from two different distributions: $\mathcal{D}_{aug_1}^{G}$ and $\mathcal{D}_{aug_2}^{G}$. Given an input image $\im$, the set of augmentations $\mathcal{V}=\{\aug^1, \aug^2, \cdots \aug^{N_L+2}\}$ is generated following \Cref{alg:spade-multi}. Lines 4-7 from \Cref{alg:savit} can be replaced with all lines from \Cref{alg:spade-multi} to make it implement the multi-crop strategy. $\aug^1$ and $\aug^2$ correspond to the global-crops while $\aug^3, \aug^4, \cdots, \aug^{2+N_L}$ correspond to the local-crops. More details can be found in the code.

\subsection{Loss expression with Multi-Crop}
The loss expression for $\mathcal{L}_G$ and $\mathcal{L}_L^{sim/geo}$ are slightly affected due to the multi-crop strategy. Only the 2 global-crops are fed to the teacher while the student is fed all crops in $\mathcal{V}$. The definition of the global- and local-representation set are thus slightly changed compared to \Cref{sec:loss}. Given a student backbone $f_s$ and teacher backbone $f_t$ as well as a set $\mathcal{V} = \{\aug^1, \aug^2, \cdots\, \aug^{2+N_L}\}$ containing $2+N_L=|\mathcal{V}|$ augmented views of the same input image, a forward pass of $\aug^1$ and $\aug^2$ in the teacher network and a forward pass of all augmentations $\aug^n$ in the student network results in:
\begin{enumerate}
    \item two sets of global\footnote{The local/global terminology used here should not be confused with the local/global terminology of the multi-crop strategy. We refer the reader to \Cref{sec:globalvslocal} for more information on global- and local-representations.}-representations $\bar{\mathcal{Z}}_s = \{\bar{f_s}(\aug) : \aug \in \mathcal{V}\}$ and \\ $\bar{\mathcal{Z}}_t = \{\bar{f_t}(\aug^1), \bar{f_t}(\aug^2)\}$\\
    \item two sets of local-representations ${\mathcal{Z}}_s = \{{f_s}(\aug) : \aug \in \mathcal{V}\}$ and \\ ${\mathcal{Z}}_t = \{{f_t}(\aug^1), {f_t}(\aug^2)\}$\\
\end{enumerate}

Given the new definition of the above sets, the only changes in \cref{eq:global_loss}, \cref{eq:total_similarity_loss} and \cref{eq:total_geometric_loss} are the normalization constants. \Cref{eq:global_loss} becomes
\begin{equation}
    \mathcal{L}_G = \frac{1}{2(N_L+1)} \sum_{\glob \in \bar{\mathcal{Z}}_t} \sum_{\substack{\glob' \in \bar{\mathcal{Z}}_s \\ \aug \neq \aug'}} H\left(\bar{h}(\glob), \bar{h}(\glob')\right)
\end{equation}
\Cref{eq:total_similarity_loss} becomes
\begin{equation}
    \mathcal{L}_{L}^{sim} = \frac{1}{2(N_L+1)} \sum_{\rep \in \mathcal{Z}_t} \sum_{\substack{\rep' \in \mathcal{Z}_s \\ \aug \neq \aug'}} L_L^{sim}(\dense, \dense')
\end{equation}
and \Cref{eq:total_geometric_loss} becomes
\begin{equation}
    \mathcal{L}_{L}^{geo} = \frac{1}{2(N_L+1)} \sum_{\rep \in \mathcal{Z}_t} \sum_{\substack{\rep' \in \mathcal{Z}_s \\ \aug \neq \aug'}} L_L^{geo}(\dense, \dense')
\end{equation}

\begin{figure}[t]
  \centering
  \includegraphics[width=\linewidth]{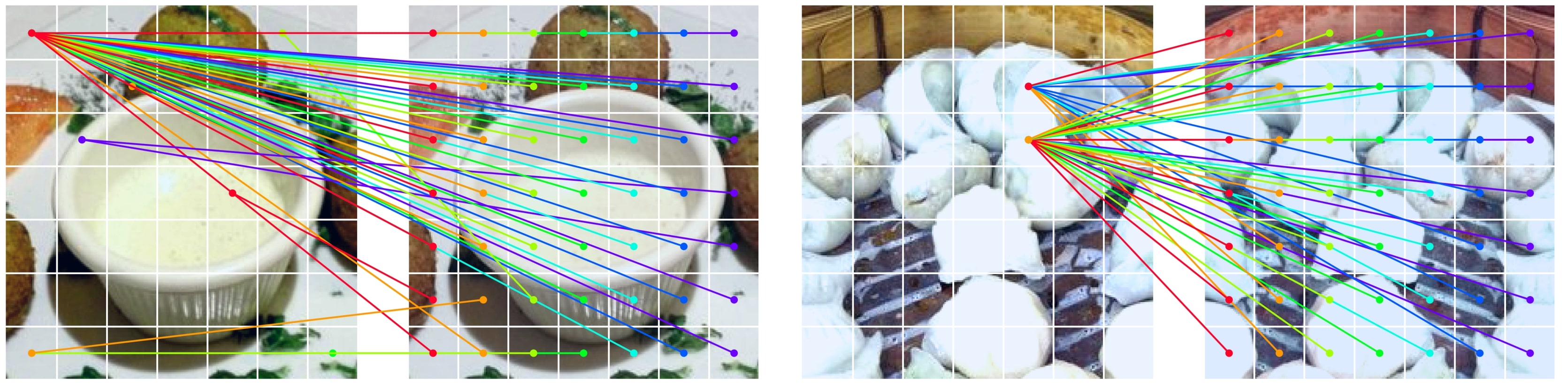}

  \caption{\textbf{Visualization of the collapse of the similarity matching function.} This example shows the matchings of the \texttt{Similarity} setting trained on Food-101.}
  \label{fig:collapse_esvit}
\end{figure}

\begin{figure*}[p]
\centering
\includegraphics[width=0.9\textwidth]{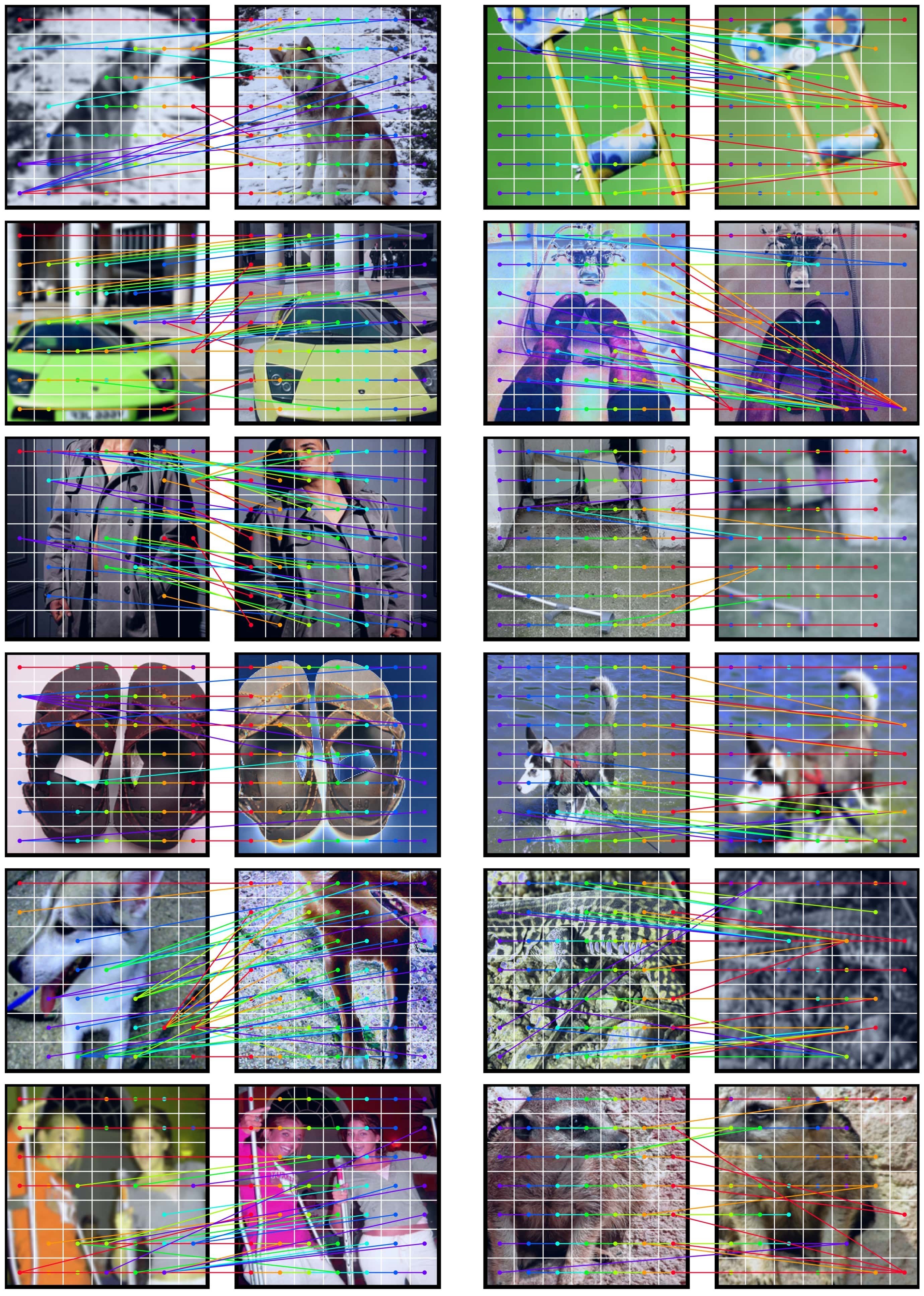}
\caption{\textbf{Visualization of the training matchings between 2 augmentations (both are global-crops) of an input image with the \texttt{Similarity} setting.} Each location on a coarse grained grid (corresponding to the area of an output token) on the right view is linked to the best matching location on the left view based on their similarity. Colors are used only to better differentiate different matchings. (extended version of the left part of \cref{fig:comparison_matchings} in the main paper)}
\label{fig:global-global-similarity}
\end{figure*}
\begin{figure*}[p]
\centering
\includegraphics[width=0.9\textwidth]{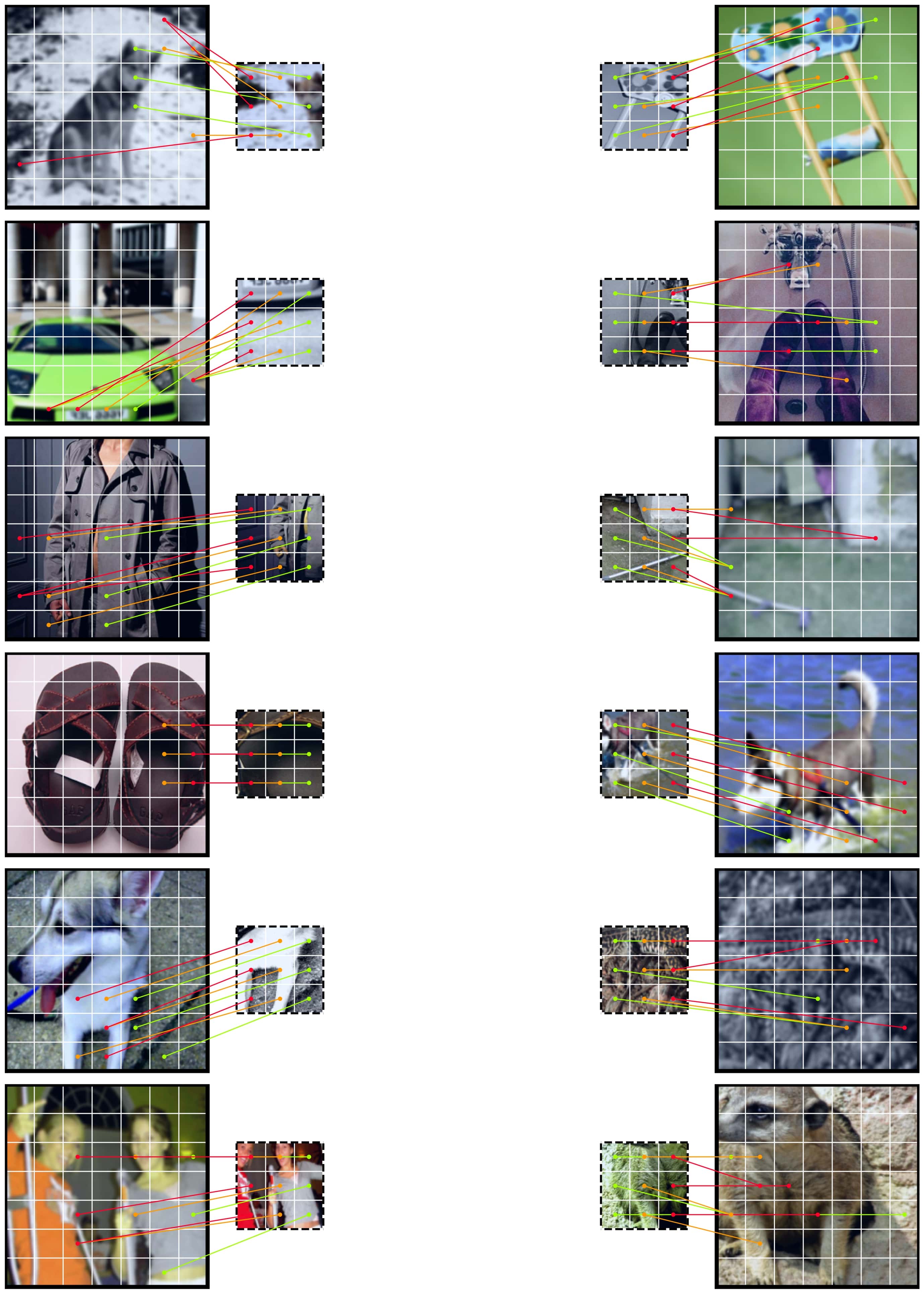}
\caption{\textbf{Visualization of the training matchings between 2 augmentations (1 global- and 1 local-crop) of an input image with the \texttt{Similarity} setting.} Each location on a coarse grained grid (corresponding to the area of an output token) on the right view is linked to the best matching location on the left view based on their similarity. Colors are used only to better differentiate different matchings.}
\label{fig:global-local-similarity}
\end{figure*}
\begin{figure*}[p]
\centering
\includegraphics[width=0.9\textwidth]{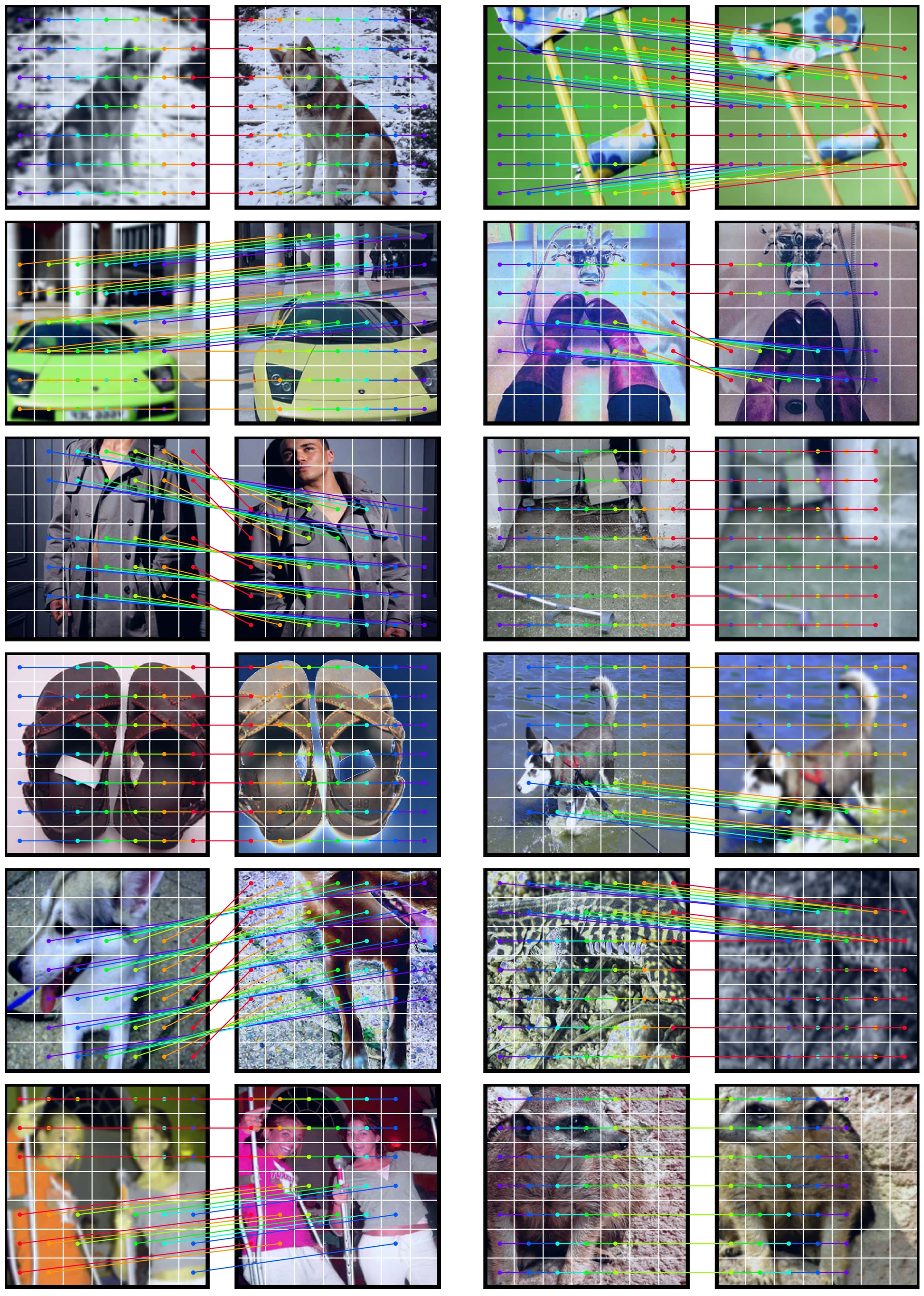}
\caption{\textbf{Visualization of the training matchings between 2 augmentations (both are global-crops) of an input image with the \texttt{Geometric} setting.} Each location on a coarse grained grid (corresponding to the area of an output token) on the right view is linked to the best matching location on the left view based on their similarity. Colors are used only to better differentiate different matchings. (extended version of the right part of \cref{fig:comparison_matchings} in the main paper)}
\label{fig:global-global-geometric}
\end{figure*}
\begin{figure*}[p]
\centering
\includegraphics[width=0.9\textwidth]{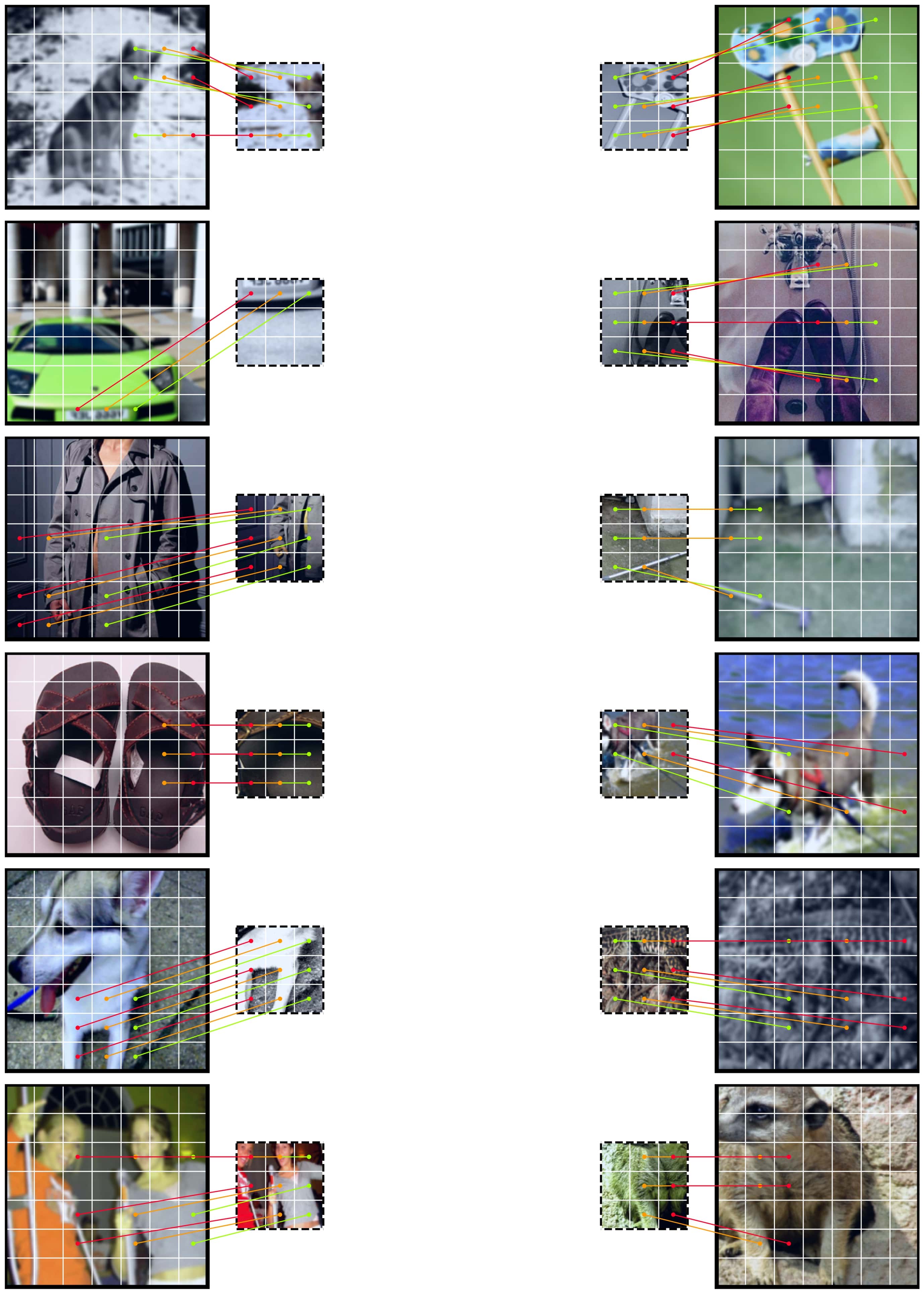}
\caption{\textbf{Visualization of the training matchings between 2 augmentations (1 global- and 1 local-crop) of an input image with the \texttt{Geometric} setting.} Each location on a coarse grained grid (corresponding to the area of an output token) on the right view is linked to the best matching location on the left view based on their similarity. Colors are used only to better differentiate different matchings.}
\label{fig:global-local-geometric}
\end{figure*}
% \clearpage
% }

\begin{figure*}[t]
\centering
\includegraphics[width=0.96\textwidth]{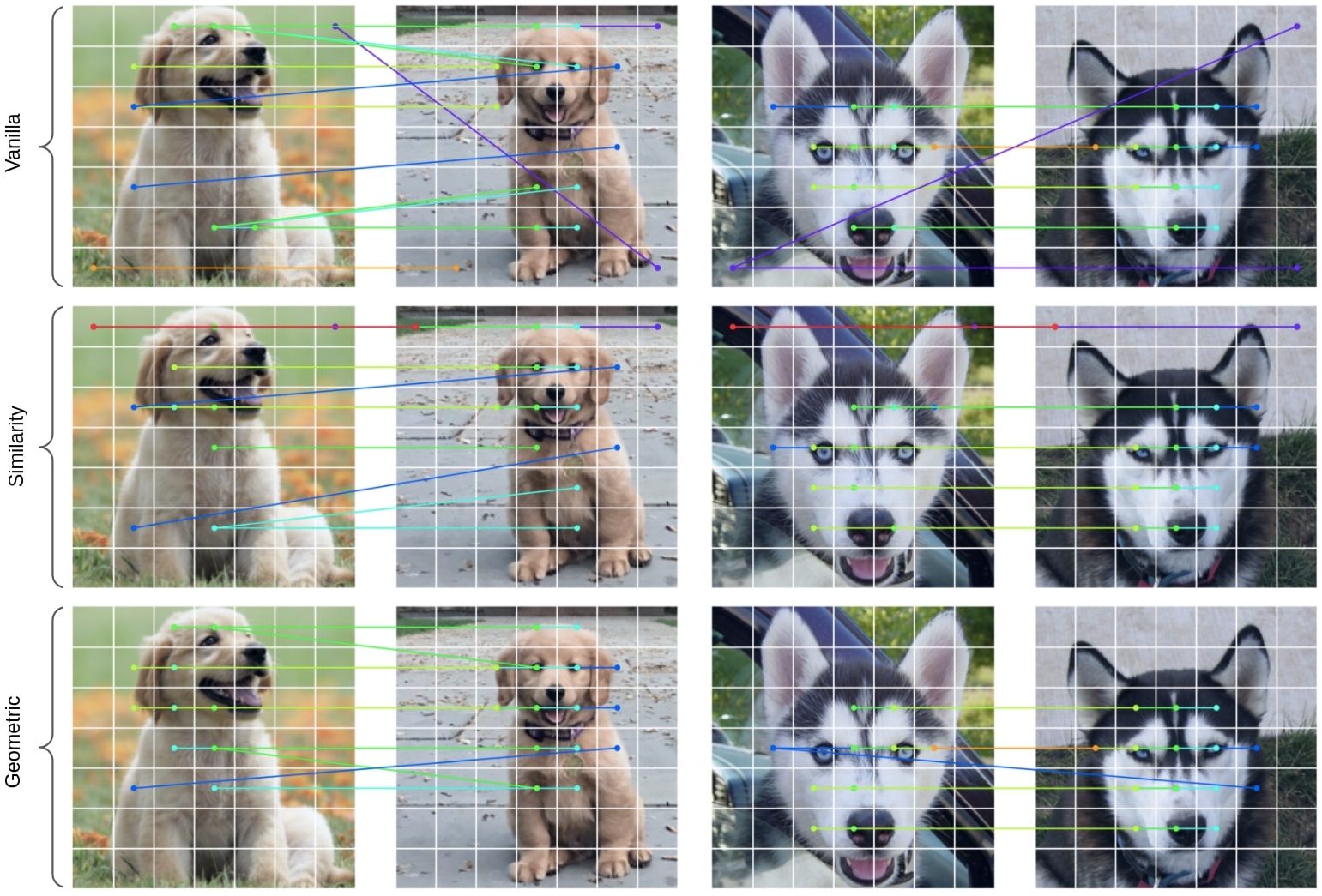}
\caption{\textbf{Visualization of the correspondence mapping between 2 different images.} The top 15 matchings are shown for all 3 settings. The matchings are obtained based on the similarity of the learned local-representations. Colors are used only to better differentiate different matchings.}
\label{fig:dog-dog}
\end{figure*}
% \clearpage
% }

\section{Training matchings visualized}
To build a better intuition of the coherence that the local-representation loss enforces, we illustrate a few pairs of augmentations $\aug$ and $ \aug'$ and show how the local-representations are matched during the training phase both for the \texttt{Similarity} (\cref{fig:global-global-similarity} and \cref{fig:global-local-similarity}) and \texttt{Geometric} setting (\cref{fig:global-global-geometric} and \cref{fig:global-local-geometric}). This is done both for a pair of 2 global-crops as well as a pair of 1 global-crop and 1 local-crop. Using a Swin-T/W=7 \cite{swin} backbone results in dense-representations which are downscaled with a factor of 32 compared to the augmentations. Global-crops of size $224\times224$ result in a dense-representation of spatial dimension $7\times7$ while local-crops of size $96\times96$ result in a dense-representation of spatial dimension $3\times3$.

As mentioned in the paper, the training matchings of the \texttt{Similarity} setting depend on the state of the local-representations. The visualization here use a network trained until the last epoch (300) on ImageNet-1k. The images shown are from the validation set of ImageNet-1k. The matchings from \Cref{fig:global-global-geometric} (\texttt{Geometric} setting) seem to be more well behaved than the matchings from \Cref{fig:global-global-similarity} (\texttt{Similarity} setting). Moreover, we can observe cases of collapse of the similarity based matching function in \Cref{fig:global-global-similarity} even though the network is trained on a large scale dataset (ImageNet-1k). This happens when the photometric transforms applied to both crops are highly different from each other (e.g. dog in 5th row of \Cref{fig:global-global-similarity}).

\section{Collapse of the similarity matching function visualization}
The collapse of the similarity matching function when trained on Food-101 \cite{food101} can be visualized in \Cref{fig:collapse_esvit}. Similar effects (though less strong) can be observed in \Cref{fig:global-global-similarity} (trained on ImageNet-1k \cite{imagenet}).

\section{Correspondence mapping between 2 different images}
Although not the goal of our work, we show qualitative results of the correspondence mapping between two different images. We use 2 pairs of 2 images with similar semantics and show a visual alignment between the two. Local-representations from each image are linked to the most similar local-representation in the other image based on their cosine similarity. The 15 assignments with the highest similarities are shown in \Cref{fig:dog-dog}. This is done with Swin-T backbones trained on all 3 different settings (\texttt{Vanilla}, \texttt{Similarity} and \texttt{Geometric}). Overall, all settings seem to show decent alignments of the 2 images though qualitative gains (on the correspondence mapping) can be observed with the additional local cues, especially in the \texttt{Geometric} setting.

\end{document}